 \documentclass[final,1p,times,authoryear]{elsarticle}

\usepackage{amssymb}
\usepackage{lipsum}

\usepackage{amssymb}
\usepackage{graphicx}
\usepackage{subfig}
\usepackage{xcolor}
\usepackage{amsmath} 
\usepackage{multirow}
\usepackage{adjustbox}
\usepackage{amsfonts}
\usepackage{wrapfig}
\usepackage{subcaption}
\usepackage{amsmath,amsfonts}
\usepackage{algorithmic}
\usepackage{algorithm}
\usepackage{array}
\usepackage{comment}
\usepackage{xr}
\usepackage{xr-hyper}
\usepackage{hyperref, cleveref}
\usepackage{enumitem}

\begin{document}

\begin{frontmatter}



\title{UnCapsTSR: An Unsupervised Transformer-based Image Super-Resolution Approach for Capsule Endoscopy Images}


\author[first]{Anjali Sarvaiya}
\author[first]{Shubh Kawa}
\author[first]{Lalit Agrawal}
\author[first]{Jagrit Joshi}
\author[first]{Kishor Upla}
\author[second]{Kiran Raja}

\affiliation[first]{
    organization={Department of Electronics Engineering, Sardar Vallabhbhai National Institute of Technology},
    city={Surat},
    country={India}
}

\affiliation[second]{
    organization={Norwegian University of Science and Technology (NTNU)},
    country={Norway}
}



\begin{abstract}
Wireless Capsule Endoscopy (WCE) captures and streams video while passing through patient's Gastrointestinal (GI) tract which is then used to examine its irregularities. While this technology is advantageous over conventional endoscopy, WCE suffers from capsule's size and wireless transmission that result in images with coarser resolution. Improving the resolution of WCE images in off-line mode (i.e., Super-Resolution (SR)) can help expert physicians to visualize details and help in diagnosis. This work presents a new approach referred as \emph{UnCapsTSR} to improve the spatial resolution of Low-Resolution (LR) WCE images making use of unsupervised training in transformer-based Generative Adversarial Network (GAN) framework. The novel design of the proposed method accomplishes the SR task without the explicit use of degradation estimation of real-world LR data and eliminates the need for true LR-HR pair. \emph{UnCapsTSR} employs a new Bilateral Total Variation (BTV) loss to ensure the spatial continuity in the SR images. By combining this loss with other GAN losses, the perceptual and quantitative qualities of the SR results is ensured.  In addition, a newly curated dataset from Kvasir capsule set is also presented in this work which is used to train the SR models of WCE images. Consequently, the generalizability of the proposed method is validated by conducting numerous experiments on different WCE datasets i.e., KID and GIANA datasets whose samples are not present in the training process. A new non-reference metric called Endoscopy Quality Metric (EndoQM) is also presented to ensure the quantitative performance of SR results of domain specific WCE data. The experimental analysis demonstrates the consistent improvement over the other state-of-the-art unsupervised SR approaches, both in terms of qualitative and quantitative evaluations on different no-reference quality measures such as NIQE, BRISQUE and PIQE in addition to EndoQM. Along with statistical evaluation the \emph{UnCapsTSR} gains 40 to 80 \% significant quantitative improvement in terms of EndoQM from LR to SR for all above datasets. \textcolor{blue}{Code link:https://github.com/KawaShubh/UnCapsTSR/tree/main}
\end{abstract}




\end{frontmatter}




\section{Introduction}
\label{introduction}
\begin{figure*}[t!]
    \centering
    \subfloat[\centering LR \\ (9.4735)]{{\includegraphics[width=3.2cm,height=3.2cm]{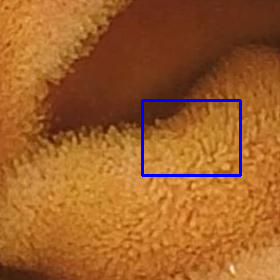} }}%
    \subfloat[\centering SRResCGAN \citep{SRResCGAN} \\ (10.0728)]{{\includegraphics[width=3.2cm,height=3.2cm]{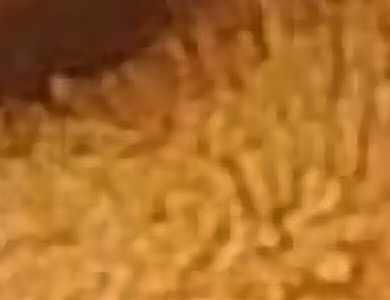} }}%
    \subfloat[\centering DUSGAN \citep{dusgan} \\ (14.0524)]{{\includegraphics[width=3.2cm,height=3.2cm]{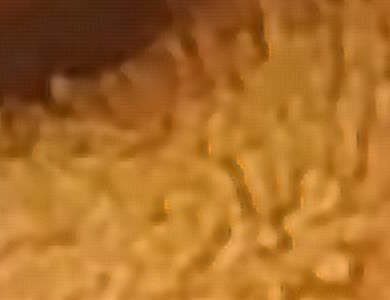} }}%
    \subfloat[\centering dSRVAE \citep{dSRVAE} \\ (7.5483)]{{\includegraphics[width=3.2cm,height=3.2cm]{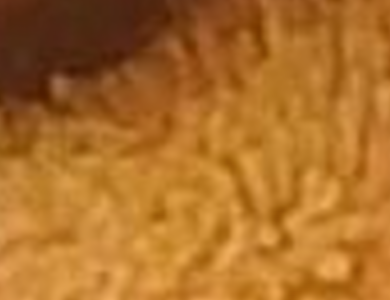} }}\\
    \vspace{-0.3cm}
    \subfloat[\centering ZSSR \citep{zssr} \\  (12.9962)]{{\includegraphics[width=3.2cm,height=3.2cm]{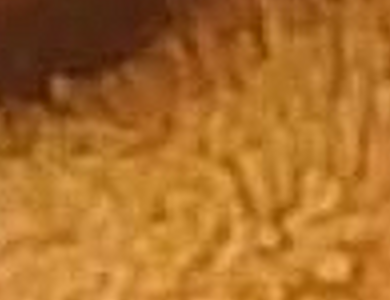} }}%
    \subfloat[\centering DASR \citep{arr41} \\  (14.3810)]{{\includegraphics[width=3.2cm,height=3.2cm]{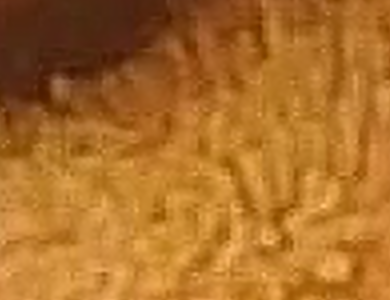} }}%
    \subfloat[\centering MDASR \citep{MDASR} \\ (7.8093)]{{\includegraphics[width=3.2cm,height=3.2cm]{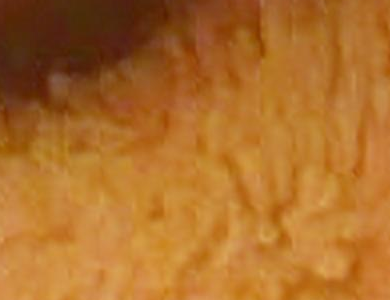} }}%
    \subfloat[\centering Proposed  \\(EndoQM$\downarrow$ = 6.8817)]{{\includegraphics[width=3.2cm,height=3.2cm]{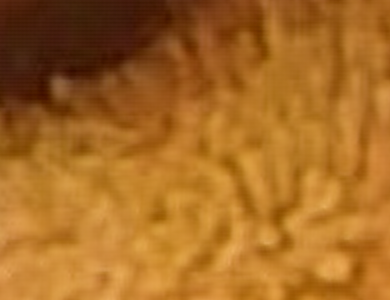} }}%
    \caption{The illustration of SR performance of the proposed method (i.e., UnCapsTSR) on WCE LR image in comparison to other state-of-the-art methods. The values of newly proposed EndoQM measure is also indicated alongside of each SR result in the bracket.}%
    \label{First_image}%
    \vspace{-0.4cm}
\end{figure*}

Computerized Tomography (CT), Positron Emission Tomography (PET), Magnetic Resonance Imaging (MRI), ultrasound imaging, endoscopy ad alike imaging approaches provide visual representation of interior parts of human body that are essential for clinical \& medical analysis and intervention. Wireless Capsule Endoscopy (WCE) is one such imaging approach that is used to examine Gastrointestinal (GI) abnormalities. This technology is comparatively painless compared to the conventional endoscope based examinations which is often painful and stressful \citep{2002}. It makes use of a pill size imaging device typically in length and diameter of $26$ $mm$ and $11$ $mm$, respectively equipped with an optical dome, illuminator, imaging sensor, battery, and RF transmitter that can be easily ingested by patients and help in diagnosis of GI tract abnormalities \citep{WCE2000}. Due to the limited size of hardware with limited battery  and low field of view of CMOS camera, WCE often hinders acquiring High-Resolution (HR) images/videos. While HR images are invariably preferred in all vision-driven applications, LR images are obtained in WCE with limited high-frequency details making diagnosis challenging \citep{2002}. Additionally, given the limited battery size of pill camera, WCE captures images with coarser resolution with compression. The motion of pill camera also results in images subjected to different artifacts such as motion blur, bubbles, specular reflections, floating objects and pixel saturation which need to be fixed by post-procedural corrections to avoid performance deviation in diagnosis \citep{Swainiv48}. While one can enhance resolution by increasing the size of the optics and sensor array, it is not always an easy solution due to reasonable cost and pill size considerations. To circumvent the aforementioned limitations in other applications, software-driven techniques referred as Super-Resolution (SR) is used for the reconstruction of missing details in the LR observations by posing it as an inverse problem \citep{SISRWCE}. Thus, the Medical SR (MedSR) is used to improve visual appearance of anatomical structures to help precise and reliable medical diagnosis.

Several SR techniques have been proposed for traditional RGB images \citep{NN, bilinear, bicubic, SRCNN, VDSRN, RCAN}. Early methods using CNNs (e.g., SRCNN \citep{SRCNN}) have been improved with Transformer-based methods (e.g., TTSR \citep{TTSR}) that leverage self-attention mechanisms to model long-range dependencies more effectively. Unlike CNNs, which are biased toward local spatial structures, Transformers excel in capturing global context, making them particularly useful for medical imaging. Their scalability, robustness to corruption, and ability to handle large datasets have driven significant interest in Transformer-based models for medical imaging applications.

Despite significant progress in SR algorithms, their application to WCE images remains relatively underexplored. Further, traditional SR approaches for regular RGB images employ supervised training, where models learn to upscale LR images by leveraging paired LR and HR images. The availability of HR endoscopy images is limited, and acquiring well-aligned LR-HR image pairs in the medical domain is particularly challenging. In some cases, synthetic generation of these LR-HR pairs is considered, but this approach is only feasible when the image acquisition process is meticulously controlled and well-defined. Typically, only an approximate understanding of the acquisition process is available, and LR images are often simulated using known downsampling techniques, such as bicubic downsampling. Consequently, supervised SR models are trained on these synthetic LR-HR pairs, where the LR images do not correspond to true camera-captured observations. As a result, such models tend to learn the degradation patterns of simulated LR samples, limiting their ability to generalize to real-world, clinically obtained LR images. The challenge is further exacerbated in medical imaging, where generating true LR-HR pairs by manipulating camera settings is impractical and often infeasible. For these reasons, we propose a deep learning architecture that is trained in an unsupervised manner, eliminating the need for precisely aligned LR-HR image pairs. This unsupervised training approach allows the model to learn directly from the available data without relying on synthetic pairings, thus improving its generalization performance when applied to real-world WCE images. Further, to gain deeper insights into the distributional disparities among different modalities such as natural, conventional endoscopy and WCE images we employ t-distributed Stochastic Neighbor Embedding (t-SNE) visualization to examine their feature space in Fig. 1(a) and Kernel Density Estimation (KDE) plots of WCE images and natural images to investigate the pixel intensity distributions in Fig. 1(b) in \emph{supplementary material}.

This work proposes \emph{UnCapsTSR}, an unsupervised approach for image super-resolution of WCE images using Transformer without explicit need for true LR-HR pair as it is impossible to obtain such data in WCE. The proposed SR model i.e., \emph{UnCapsTSR} employs a Generative Adversarial Network (GAN) architecture with a transformer-based generator and dual discriminators. Unlike traditional GAN based SR approaches, the proposed model leverages the power of transformers to capture global and local dependencies effectively, while the dual-discriminator framework ensures enhanced reconstruction quality and realism. The incorporated Bilateral Total Variation (BTV) loss preserves spatial continuity while penalizing inconsistencies. It smoothens SR image across multiple spatially shifted pixel neighborhoods and hence, effectively reduces artifacts enhancing texture fidelity leading to improved perceptual quality in the reconstructed SR images. A sample of SR performance obtained using the proposed method (i.e., \emph{UnCapsTSR}) is highlighted in Fig.~\ref{First_image} in comparison to other state-of-the-art methods on an LR image of Kvasir dataset along EndoQM score for each SR image. While eliminating the need for explicit LR-HR pair in a true sense of unsupervised setting, this work results in following key contributions:

\begin{itemize}[leftmargin=*]
    \item \textbf{Unsupervised training:} Unlike the supervised SR methods exhibiting ineffective generalizability to the real WCE samples, the proposed model i.e., \emph{UnCapsTSR} employs transformer-based GAN architecture for SR of WCE images for upscaling factor $\times 4$ which is trained in unsupervised manner, addressing the unavailability of paired LR-HR clinical datasets.
    
    \item \textbf{Transformer-based architecture design:} In \emph{UnCapsTSR}, a novel GAN architecture with a transformer-based generator and a dual-discriminator framework enhances SR for WCE images without explicit degradation estimation. The generator employs self-attention to capture global and local dependencies, while the dual discriminators refine perceptual quality. One discriminator focuses on high-frequency details like edges and textures, preserving diagnostically critical features, while the other ensures structural and perceptual consistency. This balanced approach improves both local and global quality, producing SR outputs that are visually coherent and clinically reliable for endoscopic imaging.

    \item \textbf{New BTV Loss:} Further, a Bilateral Total Variation (BTV) loss is proposed, integrated with adversarial and texture losses to preserve spatial continuity while improving perceptual quality in SR images. Unlike conventional Total Variation (TV) loss, this loss ensures the spatial smoothness  balanced with the retention of edge sharpness, which is critical for maintaining  significant details in medical imaging. Thus, by combining BTV loss with adversarial and texture losses, is particularly beneficial in medical applications, where preserving the continuity of anatomical structures alongside detailed textures is essential for accurate diagnosis. Hence, the proposed BTV loss guarantees that the SR outputs not only achieve high visual fidelity but also comply with the rigorous clinical standards.
    
    \item \textbf{New non-reference metric:} A reference-less metric Endoscopy Quality Metric (EndoQM) is proposed to gauge the SR performance of the different methods on WCE images. This metric is exclusively trained on conventional endoscopic images to assess the preservation and enhancement of edge details essential for diagnostic accuracy in SR images. Unlike other non-reference metrics such as BRISQUE, NIQE, and PIQE, which primarily assess the perceptual quality or statistical characteristics of natural images, this new metric is designed to address the unique medical imaging requirements. Thus, by prioritizing edge preservation, this metric reflects the ability of SR methods to reconstruct fine-grained details such as mucosal patterns, vascular structures, and lesions, which are essential for reliable clinical assessments in WCE images. 
    \item \textbf{New SR Dataset:} An another significant contribution through this work is the creation of a new derivative dataset specifically designed for SR task. It is curated from the publicly available Kvasir Capsule Endoscopy Dataset \citep{kvasir}. However, the original dataset is not directly suitable for SR tasks due to the presence of redundant and low-quality images, as well as unwanted border pixels that do not contribute meaningful information for training. The original Kvasir dataset is therefore curated through a meticulous pre-processing pipeline. Redundant and non-informative images were identified and removed. Additionally, unwanted border pixels were manually cropped out to improve the overall quality and relevance of the dataset for SR model training. Thus, this newly created derivative dataset provides a unique and valuable resource for advancing SR research in the medical domain, particularly for WCE images and its improved quality and relevance make it well-suited for training and evaluating SR models. 
    \item \textbf{Generalizability:} Finally, the proposed model i.e.,  \emph{UnCapsTSR} is trained on the newly derived Kvasir dataset for SR task, and for fair comparison with other state-of-the-art SR methods, we have trained all those methods on the above dataset. Further, to show the generalizability of the proposed method, all SR methods extensively evaluated on other datasets which are not part of training i.e., KID \citep{KID} and GIANA \citep{GIANA} datasets. It is evident from those experiments that, for real LR samples, the proposed SR method performs better both perceptually and quantitatively than the other SR methods in terms of numerous non-reference metrics, including BRISQUE, NIQE, PIQE, and EndoQM.

\end{itemize}
\vspace{-0.5cm}
\section{Background and Related Works}
\label{sec:relwork}

  

\subsection{Supervised SR approaches}

Prior to the emergence of deep learning, Single Image Super-Resolution (SISR) techniques predominantly relied on traditional methods, including interpolation- and reconstruction-based approaches. However, these methods exhibited limitations in handling noise, aliasing, and blurring effects, and were ineffective at higher scale factors, often failing to preserve spectral information \citep{application}. Learning-based techniques addressed some of these challenges by utilizing external data to model the relationship between low- and high-resolution images, but their performance remained constrained. The advent of deep learning has revolutionized SISR, enabling the development of advanced methods ranging from early CNNs to state-of-the-art GANs and Transformer-based approaches, which achieve significantly superior results by capturing complex image features and long-range dependencies. A comprehensive survey of supervised SR approaches is presented in the Section~{\color{red}II} in \emph{supplementary material}.

\subsection{Unsupervised SR approaches}
CNN and GAN based SR models typically follow supervised training where LR sampled are synthesized with known downsampling (i.e., bicubic) where a pair of artificial LR sample along with true HR data is used for the training SR models. Hence, a network trained with such data is prone to learn the distribution characteristics of synthetic LR samples far from that of true LR observations with different degradations making them less generalizable. While using real LR-HR pairs for training is a potential solution, it is impractical in WCE as patients cannot undergo multiple investigations for the same condition.

\cite{lugmayr2020ntire} introduced an unsupervised learning framework for training CNN-based super-resolution models on natural images without requiring paired LR-HR data. They employed CycleGAN to model the degradation process from HR to LR, enabling the SR network to be trained in a supervised fashion using synthetically degraded LR images. SRResCGAN \citep{SRResCGAN} was further proposed as an extension of the same idea using different losses such as L1, Total-Variation (TV) and VGG losses. dSRVAE \citep{dSRVAE} implements sequential de-noising followed by SR task with an encoder-decoder architecture to eliminate noise from the LR images. Further, \cite{zssr} used zero shot learning making use of internal statistics of a single image for super-resolution, eliminating the need for external training data. However, its reliance on internal image redundancy may limit performance on images with minimal recurring structures, making it less effective for highly textured or complex medical images. \cite{arr41} proposed another approach that learns degradation representations in an unsupervised manner, allowing the model to adapt to various unknown degradation patterns. However, despite its effectiveness, the method may struggle with extreme degradations and requires careful optimization to generalize well across different imaging conditions. Furthermore, USISResNet \citep{USISResNet} employed adversarial learning within an unsupervised domain adaptation framework to transform real noisy LR images into high-quality SR outputs. Similarly, \cite{dusgan} leveraged unpaired LR-HR datasets for direct unsupervised learning, demonstrating improved SR performance without the need for paired supervision. GAN-based technique with one generator and two discriminators was used by \cite{dusgan} and SR network was trained with quality, total variation and content losses. While a number of approaches can be seen for traditional computer vision applications, a limited number of works are noted for medical imaging (CT-\citep{CT7}, MRI \citep{MRI5}). Despite the noted works targeting different modalities, the MDASR model \citep{MDASR} is the only work focusing on WCE using unsupervised learning with domain adaptation techniques. The summary of different supervised and unsupervised SR methods is highlighted in Table I \emph{in supplementary material} due to space constraint. We further note the following limitations from existing works, specifically for WCE:

\vspace{-0.1cm}
\begin{itemize}[leftmargin=*]
    \item In the medical domain, most of the SR techniques \citep{mahapatra, EndoL2h, dcan} rely on supervised training which demands a true pair of LR-HR images. However, it is often impossible to acquire such pairs which compels alternate strategies to improve the spatial resolution of medical data.    
    
    \item Further, in the absence of a genuine LR-HR pair, most of the supervised SR techniques \citep{srdensenet, RCAN, SRGAN} simulate the LR observation with known degradation (i.e., bicubic downsampling). This idea of supervised training of deep models results in poor learning capabilities to the real-world data as one can note a significant distribution shift for synthetically degraded images versus naturally observed data (see Fig. 1 in \emph{supplementary material}).
    \item Additionally, the successful SR models pre-trained on natural images \citep{arr41, dusgan, dSRVAE} often fail when applied to medical images; thus, requiring domain-specific adaptation and transfer learning \citep{NN1} (see Fig. 1 in \emph{supplementary material})
    
    \item Moreover, unlike abundant works on to natural images as noted in Table I of \emph{supplementary material}, unsupervised SR models for medical images, specifically for WCE data are still in their early stages with fewer established benchmarks and datasets i.e., \citet{MDASR}. 
   
\end{itemize}

While noting the existing works for SR, it can be seen that Transformers have not been well explored for SR in medical imaging. Transformers capture global dependencies through self-attention across all inputs, whereas CNNs primarily model local spatial relationships.
Hence, several methods have been proposed for SR of natural images using Transformer \citep{transformer,transformer1}. To restrict the scope of self-attention, some researchers use local (window) self-attention, which splits the feature maps into smaller sections \citep{trans1}. In the meanwhile, they improve the interaction between windows by using the shift mechanism \citep{trans2}, overlapping windows \citep{overlap}, or the cross-aggregation operation \citep{cross}. In comparison to earlier CNN-based techniques, these methods achieve linear complexity with regard to image size. The local design must stack a lot of blocks to create global dependencies, nevertheless, in contrast to global attention. The previously stated techniques implicitly capture global information, but they make it more difficult to model spatial dependencies - a critical component of SR. Therefore, we employ Transformers to create an image SR technique that can efficiently capture global spatial information on high-resolution images at a cheap computing cost. By providing a tailored approach for the unique challenges of WCE images, including domain discrepancies and the absence of paired high-resolution data, this paper bridges this gap for obtaining superior SR images.
\vspace{-0.5cm}
\section{Proposed Framework: \emph{UnCapsTSR}}
\label{sec:propwce1}
\begin{figure}[!t]
    \centering
  \includegraphics[width=0.99\textwidth, height=6cm]{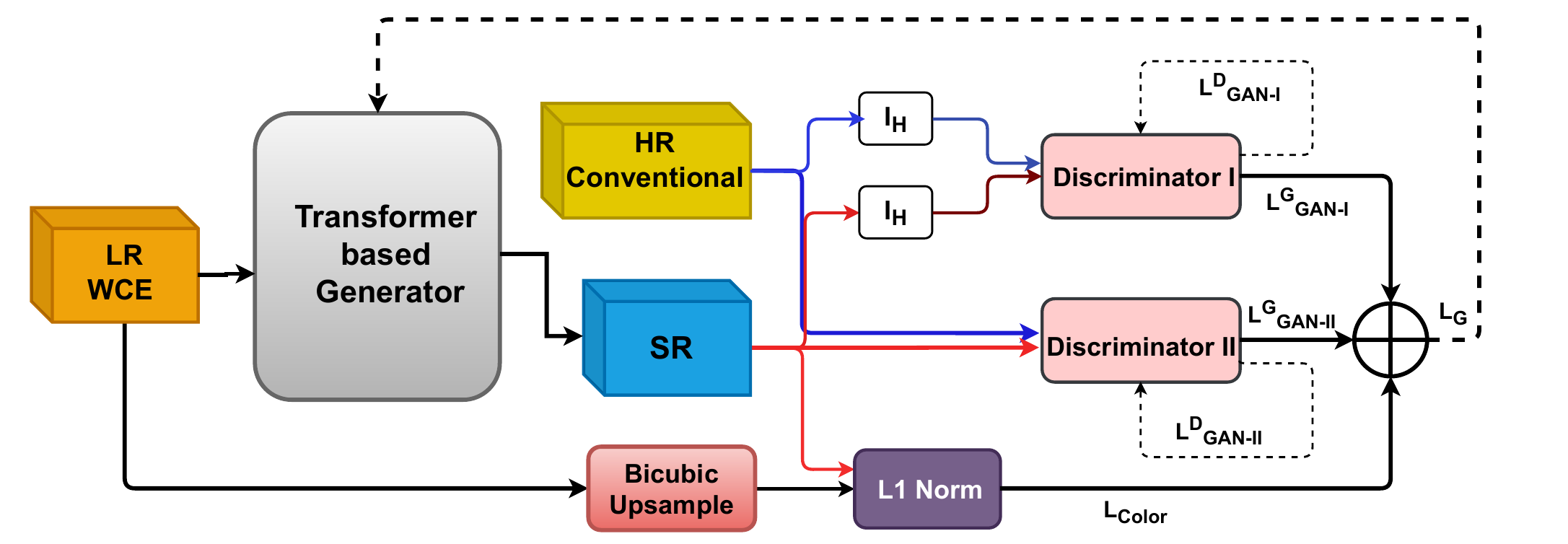}
  
    \caption{The block schematic of the proposed unsupervised SR model-\emph{UnCapsTSR} for upscaling factor of $\times4$. Here, $I_{H}$ indicates a kernel of High Pass Filter (HPF).}
    \label{fig:5}
      
\end{figure}

\begin{figure*}[t!]
    \centering
  \includegraphics[width=\textwidth]{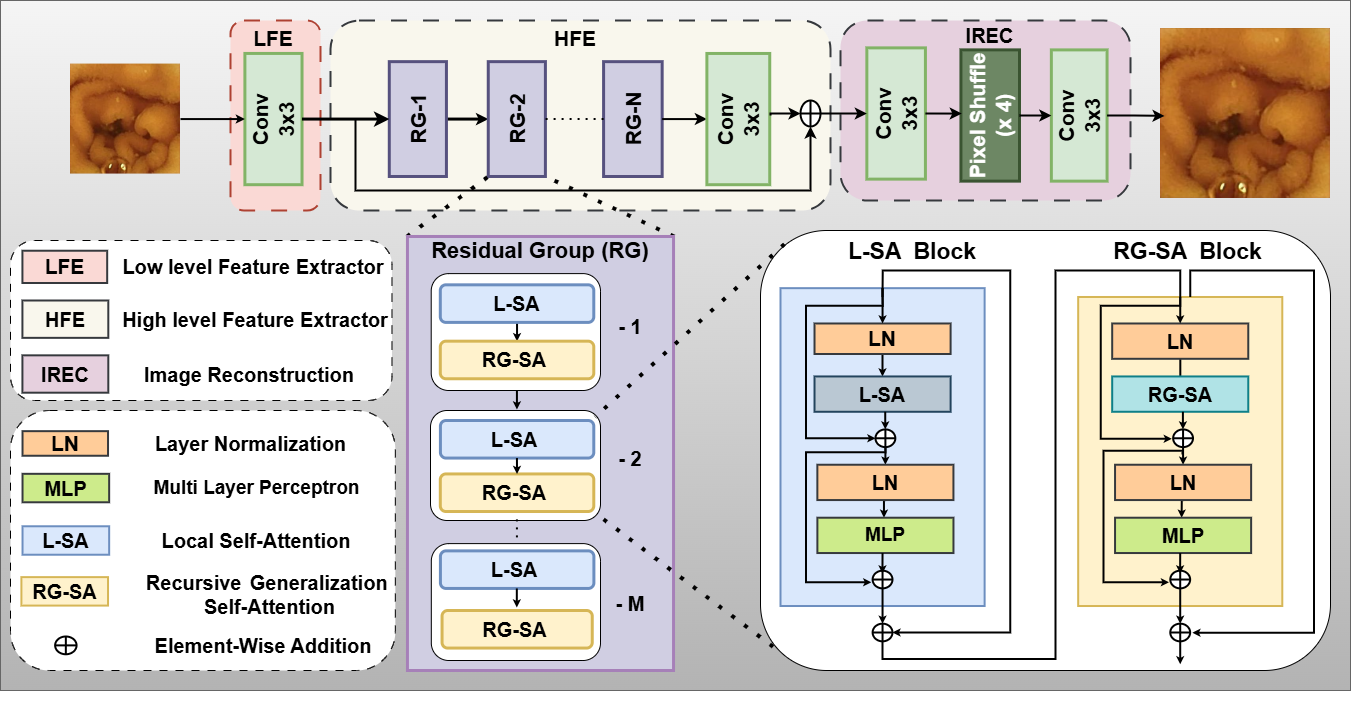}
    \vspace{-0.5cm}
    \caption{The architecture of the Transformer-based Generator network ($G$) used in the proposed method-\emph{UnCapsTSR}.}

    \label{fig:6}
    
\end{figure*}
To address the aforementioned limitations of current SR methodologies for WCE data, we propose a novel approach that utilizes a transformer model for the super-resolution employing unsupervised training for an upscaling factor of $\times4$. The block diagram of the proposed method \emph{UnCapsTSR} is depicted in Fig.~\ref{fig:5}. Our proposed approach employs transformer-based generative adversarial learning in \emph{UnCapsTSR} which incorporates three networks: a Generator (i.e., $G$), and two Discriminators (i.e., $D_I$ and $D_{II}$)  inspired by DUSGAN model \citep{dusgan} designed for natural images. However, the key differences between DUSGAN model \citep{dusgan} and the proposed model are listed below: 
\begin{itemize}
    \item \textbf{Generator Architecture:} The block diagram mentioned in \cite{dusgan} utilizes a CNN-based generator with residual-in-residual dense blocks to perform upscaling in the spatial domain. While the proposed model i.e., \emph{UnCapsTSR} replaces the CNN backbone with a transformer-based generator incorporating recursive-generalization self-attention modules to better capture long-range dependencies in WCE images. 
    \item \textbf{Loss Function Integration:} The loss function introduced in \cite{dusgan} is primarily uses color loss, adversarial losses and Quality Assessment (QA) loss. While in the proposed model we incorporate color loss, adversarial losses, texture loss and new introduced Bilateral Total Variation (BTV) loss for optimization of proposed model.\\
    \item \textbf{Domain Adaptability}: The DUSGAN \cite{dusgan} was originally designed for visible image data. Hence, the performance of such model on the medical endoscopy data exhibits degradation due to domain gap. The proposed model i.e., \emph{UnCapsTSR} explicitly addresses this limitation by leveraging domain adaptation with unpaired LR WCE and HR conventional endoscopy images, ensuring superior performance across modalities.
    \item \textbf{Multi-Modality Validation}: Unlike DUSGAN, which is limited to visible natural imagery, the \emph{UnCapsTSR} framework has been validated across multiple medical benchmarks (KID, GIANA), demonstrating its robustness and generalizability to diverse acquisition settings and clinical scenarios. Additionally, the proposed method is also validated on the other medical modality such as Retinal images. However, such finding are not presented in the DUSGAN \cite{dusgan} work. 
    \item \textbf{Quality Assessment}: The DUSGAN \cite{dusgan} reports quantitaive evaluation primarily using standard no-reference metrics such as BRISQUE, PIQE, NIQE. However, the proposed method introduces domain specific metric, Endoscopy Image Quality Metric (EndoQM), specifically trained on WCE data, to better reflect diagnostic relevance in addition to above no-reference measures. Such quantitative evaluation is not missing in the DUSGAN \cite{dusgan} paper.
\end{itemize}

The input LR image is first processed by the transformer-based generator network $G$ to produce an SR output. This generated SR image is then evaluated against an unpaired HR image by Discriminator-II ($D_{II}$), which operates directly in the SR image space. As depicted in Fig.~\ref{fig:5}, Discriminator-I ($D_I$) is employed to refine the perceptual quality of the SR image by focusing on high-frequency components from both SR and unpaired HR images. To ensure stable training of both discriminators, the framework adopts the Least Squares GAN (LSGAN) loss. To integrate unsupervised learning into the proposed framework, we introduce an additional color-consistency loss that ensures the color distribution in the super-resolved image remains consistent with that of the corresponding low-resolution input. This loss mitigates color distortions introduced during the reconstruction process and preserves the original chromatic information while enhancing spatial resolution.
Thus, an $L_1$ loss between SR and bi-cubically upsampled LR image is employed for this assignment to maintain the authenticity of color space in the SR image. In the following sub-sections, we present the detailed descriptions of each of components in above network.

\vspace{-0.3cm}
\subsection{Generator Network ($G$)}
The architecture of generator network is depicted in Fig.~\ref{fig:6}. It is based on transformer, comprising following three key elements:  (i) Low level Feature Extractor (LFE) (ii) High level Feature Extractor (HFE) (iii) Image Reconstruction (IREC).

A Low level Feature Extractor (LFE) module is utilized to capture low-frequency structural details from the Low-Resolution (LR) WCE input, denoted as $I_{LR}$. Initially, a convolutional layer with kernel of size $3 \times 3$ is applied to extract low-level feature representations. We use $64$ number of channels to extract features from the given LR observation (i.e., $I_{LR}$) to maintain complexity of the network inspired from \citet{dusgan}. Thus, the shallow features extracted by the LFE module, i.e., $F_{0}$ can be formulated as:
\begin{equation}
    F_{0} = \Phi_{LFE}(I_{LR}),
\end{equation}
where, $\Phi_{LFE}$ represents the function of LFE module.
Following this, a High level Feature Extractor (HFE) module is employed to extract high-frequency components essential for fine-detail reconstruction. This module composes a sequence of Residual Group (RG), which progressively refine feature representations, followed by a dedicated convolutional layer to enhance feature abstraction with $64$ channels. Thus, the deep feature extracted by the HFE module, i.e. $F_{d}$ can be formulated as,
\begin{equation}
    F_{d} = \Phi_{HFE}(F_{0}),
\end{equation}
where, $\Phi_{HFE}$ represents the function of HFE module. Further, to retain both low-frequency and high-frequency components, the features extracted from the LFE and HFE modules are fused via a residual connection (see Fig.~\ref{fig:6}), ensuring effective feature propagation:
\begin{equation}
    F_{fusion} = F_{0} + F_{d}, 
\end{equation}
where, $F_{fusion}$ represents the enhanced feature map obtained after residual connection. This integration preserves low-level details while enriching high-level feature representations. The fused features are then passed to the Image Reconstruction (IREC) module, which utilizes pixel-shuffle operations along with additional convolutional layers with kernel size $3 \times 3$ and with $64$ features (i.e., channel = $64$) to upsample and reconstruct the SR image. The final output is obtained as follows:
\begin{equation}
    I_{SR} = \Phi_{IREC}(F_{fusion}).
\end{equation}
where, $\Phi_{IREC}$ represents the function of Image Reconstruction module. Such  hierarchical design in \emph{UnCapsTSR} effectively combines low-level and high-level features, leading to enhanced SR results in the output.

Further, each Residual Group (RG) used in HFE module (see Fig.~\ref{fig:6}) comprises series of transformer blocks incorporated to enhance SR performance. As depicted in Fig.~\ref{fig:6}, the transformer blocks are categorized in two blocks: Local Self-Attention (L-SA) and Residual Group Self-Attention (RG-SA) blocks. These are interleaved in an alternating manner to preserve the topological structure of the module. Each transformer block consists of two Layer Normalization (LN) layers, a self-attention mechanism, and a Multilayer Perceptron (MLP) module, following the design principles of the transformer architecture. 

Such a structured configuration enables effective modeling of both local and global feature representations. The Rectangle-Window Self-Attention (Rwin-SA) proposed in \citet{cross} is applied as the Local Self-Attention (L-SA) mechanism within the transformer-based architecture. The input feature maps are partitioned into non-overlapping rectangular windows, enabling localized attention computation while maintaining a balance between computational efficiency and detailed feature representation. This mechanism allows fine-grained spatial dependencies to be captured within each window, which is particularly beneficial for preserving intricate details in WCE images.

Further, the use of Recursive-Generalization Self-Attention (RG-SA) mechanism proposed in \citet{transformer1} is to capture global contextual information efficiently with linear computational complexity. This employs a Recursive Generalization Module (RGM) to abstract the input features, irrespective of their resolution, into representative feature maps of fixed and small dimensions. This abstraction intuitively aggregates global information into compact, representative maps. Subsequently, a cross-attention mechanism is applied between the input features and representative maps, facilitating an efficient exchange of global contextual information. Given the significantly smaller size of the representative maps compared to the input features, this approach substantially reduces computational overhead. Additionally, as in the RGT \citep{transformer1}, positional encoding is not explicitly added; instead, the model relies on the windowing mechanism and recursive generalization for positional awareness. Furthermore, RG-SA incorporates an adaptive adjustment of the channel dimensions of the query, key, and value matrices within the self-attention mechanism. This adjustment mitigates redundancy in the channel domain, optimizing computational efficiency while preserving the representational power of the attention mechanism. The combination of the RG-SA with the Local Self-Attention (L-SA) in an alternate arrangement utilizes the global context in a better manner.
\subsection{Discriminator Networks (i.e., Discriminator-I \& -II)}

The proposed approach i.e.,\emph{UnCapsTSR} consists of two discriminator networks (Discriminator-I and Discriminator-II) to improve quality of SR image. Both discriminators follow the same design as presented in \citet{dusgan}. Due to page limit we have described discriminators in Section \textcolor{red}{III} in \emph{supplementary material}.

\subsection{Loss functions}
We elaborate the distinct loss functions that are associated with different blocks within the proposed unsupervised SR framework-\emph{UnCapsTSR}.
\vspace{-0.3cm}
\subsubsection{Generator:}
To improve the effectiveness of the generator network, the framework integrates a weighted combination of five distinct loss functions, as illustrated in Fig.~\ref{fig:5}.
Mathematically, the total loss (i.e., $\mathrm{L}_{G}^{}$ ) can be provided as,
\begin{equation}\label{eqn:losst}
\text{L}_G = \beta_1 L_{\text{color}} + \beta_2 L_{\text{GAN-I}} + \beta_3 L_{\text{GAN-II}} + \beta_4 L_{\text{Texture}} + \beta_5 L_{\text{BTV}},    
\end{equation}
where $\mathrm{\mathit{L}}_{color}$,
$\mathrm{\mathit{L}}_{GAN-I}$ ,
$\mathrm{\mathit{L}}_{GAN-II}$, $\mathrm{\mathit{L}}_{Texture}$, $\mathrm{\mathit{L}_{BTV}}$ reveal, color loss, generator adversarial loss with discriminator-I, generator adversarial loss with discriminator-II, texture loss and Bilateral Total Variation (BTV) loss, respectively. The weights of each loss are denoted by $\beta_i$, $i= 1,\ldots, 5$. In \emph{UnCapsTSR}, the use of unpaired LR-HR image pairs limits the applicability of traditional pixel-wise losses such as $L_1$ or $L_2$, as they fail to preserve the structural and color consistency of the LR observation in the resulting SR image. To address this limitation, a color loss (Equation~(\ref{eqn3_1})) is introduced, which computes the absolute difference between the bicubically upsampled LR image and the generated SR image.
\begin{equation}
    \label{eqn3_1}
    L_{\text{color}} = \frac{1}{P} \sum_{}^{P} 
    \left\lVert I_{SR} - B\!\left(I_{LR}\right) \right\rVert.    
\end{equation}
Here, $||\cdot||$ denotes the sum of absolute differences over all pixels in the image. This explicitly clarifies that the loss is computed across all pixel locations in the vectorized domain. $P$ stands for the size of the training batch, while $B(\cdot)$ stands for the bicubically upsampling operation. As mentioned earlier, we apply Least Squares GAN (LSGAN) loss, to enhance training stability with both discriminators. Hence, the generator loss from Discriminator-I can be represented as,
\begin{equation}
    \label{eqn 3.2}
    L_{GAN-I}^{G}=\frac{1}{P}\sum_{}^{P}\left\lVert 1-D_{I}^{}\left( I_{H}^{}\left(I_{SR}^{} \right) \right) \right\rVert.
\end{equation}
In the above equation, $D_{I}^{}$ (·) denotes the function of
discriminator-I, and $\mathrm{I}_{H}^{} $ specifies the kernel weights for a High-Pass Filter (HPF). Similarly, the equation of the adversarial generator loss from Discriminator-II is,
\begin{equation}
    \label{eqn 3.4}
    L_{GAN-II}^{G}=\frac{1}{P}\sum_{}^{P}\left\lVert 1-D_{II}^{}\left(I_{SR}^{} \right) \right\rVert,
\end{equation}
where $D_{II}^{}  (\cdot)$ indicates the function of discriminator-II. 
 
Importantly, to reconstruct the texture of SR image, we additionally propose to add texture loss \citep{textureloss}. This loss utilizes the gram matrix, where each element represents the inner product between the vectorized feature maps $\phi_{i}$ and $\phi_{j}$ at a specific layer of a pre-trained deep network. The gram matrix effectively captures the spatial correlations of feature maps, representing the global texture statistics of the image. Building upon this, the texture loss is defined as the MSE of the difference between the gram matrices of the generated super-resolved image $(I_{\text{SR}})$ and the high-resolution image $(I_{\text{HR}})$. The texture loss is stated as \citep{untexture},
\begin{equation}
    L_{\text{Texture}} = \left\| G\left(\phi(I_{\text{SR}})\right) - G\left(\phi(I_{\text{HR}})\right) \right\|_2^2.
\end{equation}
Since LR and HR images in our setup are unpaired, standard SR approaches struggle to map low-resolution inputs to a plausible high-resolution output. By enforcing similarity between the texture distributions of generated SR images and real HR images, texture loss helps in learning a domain-consistent super-resolution model that produces realistic images suitable for clinical use. We extend the applicability of the above formulation by taking the Gram-matrix statistics of HR images as domain-level priors rather than image-specific targets. Specifically, the proposed method enforces that the texture distributions of generated SR images remain consistent with the global distribution of textures observed in high-quality endoscopic images. Thus, enforcing similarity between texture distributions refers to aligning the second-order feature correlations of generated SR outputs with those typical of the HR domain. This alignment ensures that the reconstructed SR images exhibit the global texture statistics characteristic of high-quality endoscopic images, including mucosal folds, vascular textures, and surface continuity, even in the absence of exact LR-HR pairs. In support of this, authors in \cite{hu2021multi, femasr} utilized this concept, where multi-scale texture statistics in feature space are matched across domains to capture realistic textures without paired supervision. Inspired by this, our unpaired texture loss leverages Gram-matrix–based distribution alignment to reduce the domain gap between LR WCE images and HR conventional endoscopy images, thereby producing SR outputs that are perceptually and diagnostically reliable. Further, we propose the Bilateral Total Variation (BTV) Loss as a customized loss function specifically designed for SR of capsule endoscopy images, capturing both local and contextual information. This loss incorporates a weighted summation of pixel-level differences within a predefined spatial neighborhood also effectively models bilateral relationships between a central pixel and its surrounding pixels. 
\vspace{-0.3cm}
   \begin{multline}
    \label{eqn 3.3}
   \mathrm{L}_{BTV}^{}({I}_{SR}, n, \epsilon)= \frac{\lambda}{C \cdot H \cdot W}  \sum_{k=-n}^{n} \sum_{l=-n}^{n} \sqrt{({I}_{SR}^{}(i, j)- {I}_{SR}^{}(i + k, j + l))^2 + \epsilon}
\end{multline} 

In the above equation, $\lambda$ is weighing factor which is set to $1\times e^{-2}$. $C$, $ H$ and $W$ are number of channels, height and width of the image, respectively. $\epsilon$ is the regularization term for numeric stability which is set to $1\times e^{-8}$. Although the squared difference term under the square root is mathematically non-negative, in practice, very small values often occur in uniform regions of WCE images (e.g., lumen or flat mucosa). Under mixed-precision training, such small values may lead to unstable gradient computations, including division-by-zero or undefined values (NaNs). The addition of  $\epsilon$ guarantees that the square root and subsequent derivatives remain well-behaved across all regions of the image, thus improving training robustness without affecting the overall loss behavior. This loss ensures the preservation of fine-grained textures and structural details which are critical for medical analysis. By summing over neighboring pixels, the loss penalizes inconsistencies in spatial continuity, enhancing the structural integrity of the reconstructed SR images. Additionally, the inclusion of a small regularization term $\epsilon$ ensures numerical stability, while the normalization factor balances the contribution of the loss across different image scales and channels. Thus, the BTV loss is tailored to address the unique challenges of capsule endoscopy imaging, such as noise, motion artifacts, and low-resolution inputs, ultimately improving the quality and diagnostic utility of the SR images.

\subsubsection{ Discriminator-I}
To distinguish high-frequency features between SR and HR images, Discriminator-I is employed. As it aims to align the high-frequency components of the SR image with those of the unpaired HR images, the associated LSGAN loss for Discriminator-I can be formulated as follows:
\begin{equation}
    \label{eqn 8}
    L_{GAN-I}^{D}=\frac{1}{P}\sum_{}^{P}\left( \frac{| D_{I}^{}(I_{H}^{}(I_{SR}^{}))| +\left| 1-D_{I}^{}(I_{H}^{}(I_{HR}^{})) \right|}{2} \right).
\end{equation}
Here,  $D_{I}^{} $(·) represents operation of Discriminator-I.

\subsubsection{Discriminator-II}
Here, Discriminator-II is trained in an adversarial manner to distinguish the generated SR image from the original unpaired HR image. To stabilize the adversarial training process, the Least Squares GAN (LSGAN) loss is utilized for Discriminator-II, which is formulated as:
\begin{equation}
    \label{eqn 7}
    L_{GAN-II}^{D}=\frac{1}{P}\sum_{}^{P}\left( \frac{ | D_{II}^{}(I_{SR}^{})| +\left| 1-D_{II}^{}(I_{HR}^{}) \right|}{2} \right),
\end{equation}
where, $D_{II}(\cdot)$ denotes the function of Discriminator-II and $I_{HR}^{}$ is the unpaired HR image in the dataset.

\section{Proposed Quality Metric - EndoQM}
Another novel contribution of this work is the introduction of a new quality metric termed Endoscopy Quality Metric (EndoQM), which is specifically tailored for evaluating the quality of super-resolved endoscopic images. The design of EndoQM is motivated by the limitations of the widely used Natural Image Quality Evaluator (NIQE) proposed by \citet{niqe}. NIQE is a completely blind, no-reference quality assessment measure that does not require subjective opinion scores. It operates by modeling Natural Scene Statistics (NSS), extracted from high-quality natural images, and fitting them to a Multivariate Gaussian (MVG) distribution. The quality of a test image is then quantified as the distance between its NSS features and this pristine natural-image distribution. While NIQE has proven effective for generic natural image assessment, its statistical assumptions are based on natural scenes, which exhibit very different textures, illumination patterns, and structural distributions compared to medical imagery such as Wireless Capsule Endoscopy (WCE). As shown in Fig. 1(b) in \emph{supplementary material}, the statistical distribution of WCE images is narrower and more constrained, making NIQE less reliable in this domain.
To overcome this limitation, EndoQM adopts the NIQE framework but re-estimates the MVG distribution using a curated set of high-quality WCE images instead of natural images. By learning the statistical model from WCE data, EndoQM captures the modality-specific features such as mucosal textures, vascular patterns, and illumination artifacts that are diagnostically relevant. Thus, while the feature extraction and mathematical pipeline remain identical to NIQE, the reference distribution is domain-specific. This simple but crucial adaptation ensures that EndoQM aligns more closely with the clinical requirements for endoscopy, providing reliable quality evaluation even in the absence of ground-truth high-resolution references. Unlike BRISQUE, which depends on predefined natural scene statistics, or PIQE, which is overly sensitive to pixel-wise distortions, NIQE’s Multivariate Gaussian (MVG)-based feature representation allows EndoQM to effectively model the unique structural and textural attributes of endoscopic images. This adaptability ensures that EndoQM aligns more closely with the clinical requirements for assessing the perceptual quality of WCE images, providing a reliable metric for evaluating super-resolution performance in medical imaging applications.

\vspace{-0.4cm}
\section{Experimental Evaluations}
\label{sec:expwce}
A comprehensive analysis evaluating the effectiveness of the proposed model (i.e., \emph{UnCapsTSR}) by comparing it with other leading SR techniques for an upscaling factor of $\times 4$ is presented in this section using qualitative and quantitative measures. We examine representative patches from the output images of all competing models, offering a visual comparison of their reconstruction quality. On the quantitative side, performance is evaluated using widely recognized non-reference SR metrics, such as the Blind/Referenceless Image Spatial Quality Evaluator (BRISQUE) \citep{brisque}, Natural Image Quality Evaluator (NIQE) \citep{niqe}, and Perception-based Image Quality Evaluator (PIQE) \citep{piqe}, along with a newly introduced metric EndoQM, specifically designed to evaluate SR results of capsule endoscopy image, which are crucial for diagnostic accuracy in medical imaging. Additionally, we present an exhaustive ablation study on the different aspects of the proposed model such as loss functions and network configuration, demonstrating their effectiveness in the proposed model. To assess generalization, the proposed model is tested on different benchmark datasets, including the KID \citep{KID} and GIANA \citep{GIANA} datasets, showcasing its generalizability and applicability across various clinical scenarios. Additionally, we perform an Analysis of Variance (ANOVA) test, two tile Z-test and Kolmogorov–Smirnov (K–S) test to evaluate the statistical significance of the performance differences between the proposed method and other state-of-the-art approaches. The detailed description related to different experiments and its associated analysis are discussed further in the following sections.

\subsection{Datasets and training details}
The proposed model i.e., \emph{UnCapsTSR} has been trained in unsupervised manner, utilizing unpaired LR and HR images. A significant contribution of this research is the creation of new dataset derived specifically for the SR task, sourced from the existing Kvasir Capsule Endoscopy Dataset \citep{kvasir}, which comprises only Wireless Capsule Endoscopy (WCE) samples. As depicted in Table~\ref{dataset}, the original Kvasir dataset contains $47,236$ RGB images, each sized at $336 \times 336$ pixels, categorized based on various medical anomalies. We meticulously curated the dataset for the SR task by eliminating the unwanted portions such as redundant images and extraneous border pixels. Consequently, the newly edited SR dataset includes $10,000$ training images, $550$ validation images, and $1,000$ testing images, with each image resized to $280 \times 280$ pixels. The \emph{UnCapsTSR}, along with other models, has been evaluated on this new dataset for SR results. Further, these images serve as input in the LR block, as illustrated in Fig~\ref{fig:5}. Given that the model relies on unpaired LR-HR images, we incorporated an additional $10,000$ conventional endoscopy images \citep{conventional} with a resolution of $1024 \times 1024$ pixels as the HR block in Fig~\ref{fig:5}. The details of this conventional Kvasir dataset is mentioned in Table~\ref{dataset}. Thus, it is important to note that the training dataset in the proposed method does not consist of true LR-HR pairs; rather, the LR images are derived from WCE images, while the HR dataset comprises of images from conventional endoscopy resulting in a truly unpaired training dataset. Further, to assess the generalization capability of the proposed model, we also conducted tests on two additional datasets, namely KID \citep{KID} and GIANA \citep{GIANA} which are fully disjoint from training dataset (see Table~\ref{dataset}), alongside the newly curated Kvasir dataset.

\begin{table}[!t]
	\caption{The details of endoscopic image datasets.}
    \resizebox{\textwidth}{!}{
     
    \renewcommand{\arraystretch}{2.5}
		\begin{tabular}{lccccc}
			\hline
			\hline
			\multirow{2}{*}{\textbf{\huge Details}} & \multicolumn{5}{c}{\textbf{\huge Dataset Name}}                                                                                                                                                   \\ \cline{2-6} 
			& \multicolumn{1}{c}{\textbf{\begin{tabular}[c]{@{}c@{}}\huge Kvasir - Capsule \\ \huge\citep{kvasir}\end{tabular}}} & \multicolumn{1}{c}{\textbf{\begin{tabular}[c]{@{}c@{}} \huge KID\\ \huge\citep{KID}\end{tabular}}} & \multicolumn{1}{c}{\textbf{\begin{tabular}[c]{@{}c@{}}\huge GIANA \\ \huge\citep{GIANA}\end{tabular}}}  & \multicolumn{1}{c}{\textbf{\begin{tabular}[c]{@{}c@{}}\huge  Kvasir - Conventional\\ \huge\citep{conventional}\end{tabular}}} & \textbf{\begin{tabular}[c]{@{}c@{}}\huge Newly edited Kvasir capsule \\ \huge dataset for SR task\end{tabular}} \\ \hline
			\hline
			\textbf{\huge Year of publication} & \multicolumn{1}{c}{\huge 2021} & \multicolumn{1}{c}{\huge 2017} & \multicolumn{1}{c}{\huge 2017} & \multicolumn{1}{c}{\huge 2017} &
            \multicolumn{1}{c}{\huge -}\\
			\textbf{\huge Sensor} & \multicolumn{1}{c}{\huge Olympus EC-S10 endocapsule} & \multicolumn{1}{c}{\huge  MiroCam\textsuperscript{\textregistered}  capsule} & \multicolumn{1}{c}{ \huge PillCam\textsuperscript{\textregistered} SB3 }  & \multicolumn{1}{c}{\huge digital high-definition endoscopes } & \multicolumn{1}{c}{ \huge - }\\ 
			\textbf{\huge Image types}          & \multicolumn{1}{c}{\huge Capsule}         & \multicolumn{1}{c}{\huge Capsule}                                                                      & \multicolumn{1}{c}{\huge Colonoscopy and Capsule}                                                            & \multicolumn{1}{c}{\huge Gastrointestinal (conventional)}                                                                                  &  \huge Capsule                                                                      \\ 
			\textbf{\huge No. of Images}            & \multicolumn{1}{c}{\huge 47,238}                    & \multicolumn{1}{c}{\huge 2448}                                                                                   & \multicolumn{1}{c}{\huge \textgreater 2000}                                                                            & \multicolumn{1}{c}{\huge \textgreater 8000}                                                                                                        &  \Large 11550                                                                                  \\ 
			\textbf{\huge Image size}         & \multicolumn{1}{c}{\huge 336 $\times$ 336}                 & \multicolumn{1}{c}{\huge 360 $\times$  360}                                                                              & \multicolumn{1}{c}{\huge 576 $\times$  576}                                                                                    & \multicolumn{1}{c}{\huge 720 $\times$  574 to 1920 $\times$  1072}                                                                                    & \huge 280 $\times$  280                                                                              \\ 
			\textbf{\huge Source}              & \multicolumn{1}{c}{\huge https://osf.io/dv2ag/}     & \multicolumn{1}{c}{\begin{tabular}[c]{@{}c@{}} \huge https://mdss.uth.gr/\\  \huge datasets/endoscopy/kid/\end{tabular}} & \multicolumn{1}{c}{\begin{tabular}[c]{@{}c@{}}\huge https://endovissub2017-\\  \huge giana.grand-challenge.org/\end{tabular}} & \multicolumn{1}{c}{\begin{tabular}[c]{@{}c@{}} \huge https://datasets.simula.no/kvasir/\end{tabular}} & \huge -                                                                                      \\ 
			\hline
			\hline
		\end{tabular}
	 }
	\label{dataset}
    \vspace{-0.4cm}
\end{table}

All SR methods including the proposed method-\emph{UnCapsTSR} are trained on newly edited Kvasir dataset with an upscaling factor of $\times 4$.  The proposed approach is trained for batch size $32$, where each input image is randomly cropped to $64 \times 64$ size, and the total training iterations are kept to $100K$. The training patches are augmented using random horizontal flips and rotations with $90^\circ$, $180^\circ$ and $270^\circ $. The Adam optimizer is employed and the learning rate is set to $1 \times {10}^{-4}$ for generator, and $1 \times {10}^{-3}$ for discriminator which decays by half to twenty percent, forty percent, sixty percentage and eighty percentage of the total number of iterations within this training set. Additionally, the weightage to different losses in the total loss function (Equation~(\ref{eqn:losst})) i.e., $\beta_{i}$, $i=1,\ldots,5$ are set empirically to $1$, $1$, $1$, ${10}^{-4}$ and ${10}^{-2}$, respectively. The neighbor pixel in BTV loss is set to $5$ in the proposed method. The number of Residual Groups (RGs) used in generator is $4$ (i.e., $N =4$). Similarly, the number of L-SA and RG-SA in each RG is also kept to $4$ (i.e., $M = 4$). In \emph{UnCapsTSR} to kernel of HPF i.e., $I_{H}$, a Gaussian low-pass filter (LPF) with a $9 \times 9$ kernel, zero mean, and a variance of 0.8 is initially constructed and subsequently inverted to obtain the high-pass filter (HPF) weights, which are empirically determined.

 \begin{figure*}[!t]
 \centering
    \subfloat[\centering EndoQM $\downarrow$ = 3.1946 ]{{\includegraphics[width=3.2cm,height=3.2cm]{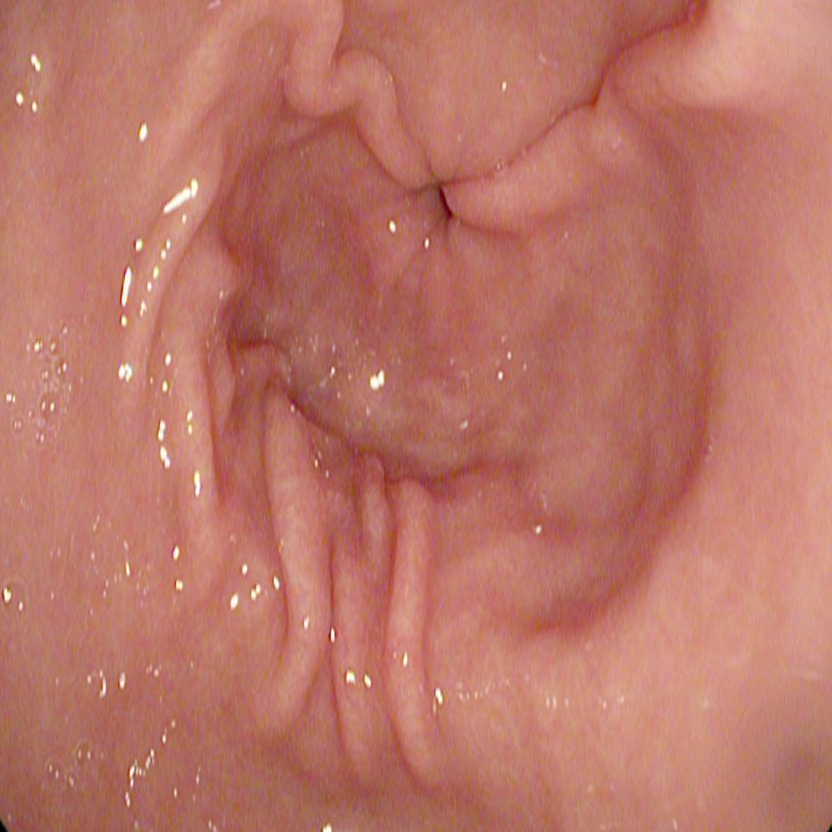} }}%
    \subfloat[\centering 9.6054 ]{{\includegraphics[width=3.2cm,height=3.2cm]{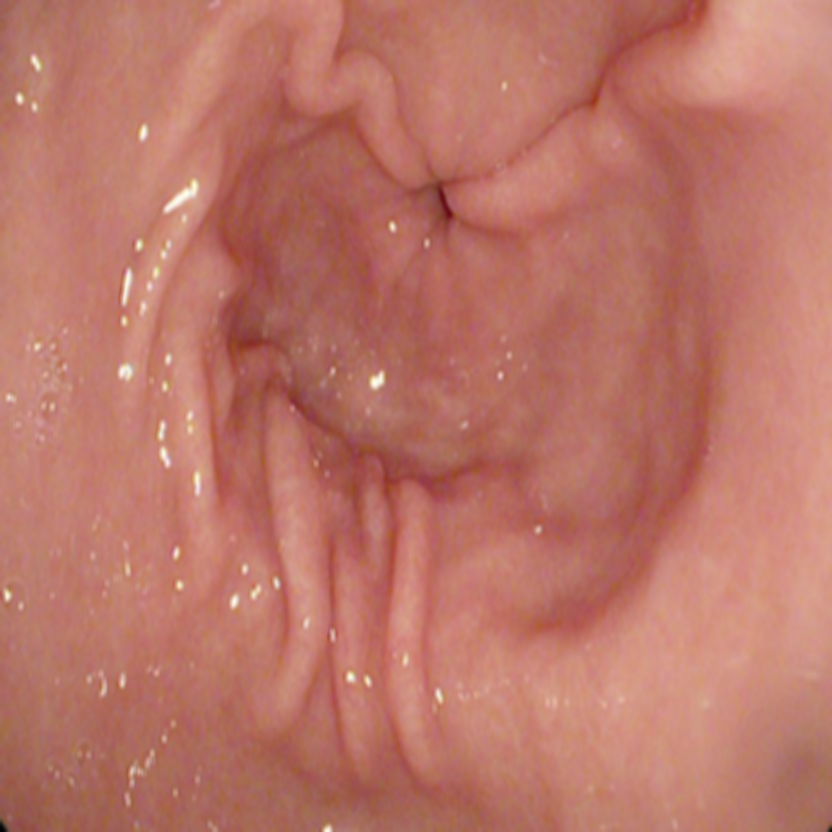} }}%
    \subfloat[\centering 10.5889 ]{{\includegraphics[width=3.2cm,height=3.2cm]{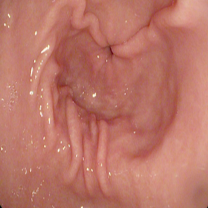} }}%
    \subfloat[\centering 10.6183]{{\includegraphics[width=3.2cm,height=3.2cm]{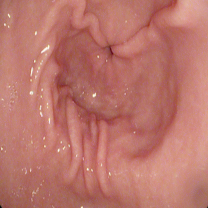} }}\\%
    
    \caption{The experimental justification showing the consistency of the newly proposed EndoQM measure through the degradation pipeline from HR to LR on a sample of conventional Kvasir dataset \citep{conventional}. Below to each result, we depict value of EndoQM metric. (a) Original image of size $832\times832$, (b) degraded version of (a) with LPF, (c) bicubically downsampled versions of (b) with dimensions $208\times208$ and (d) noisy version of (c) with size $208\times208$.}%
    \label{fig:endo1}%
    
\end{figure*}

\begin{figure*}[!t]
\centering
    \subfloat[\centering EndoQM$\downarrow$ = 7.6746 ]{{\includegraphics[width=3.2cm,height=3.2cm]{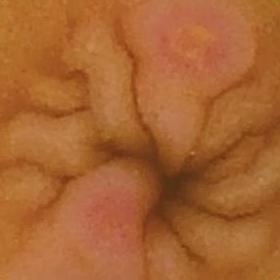} }}%
    \subfloat[\centering 15.5934 ]{{\includegraphics[width=3.2cm,height=3.2cm]{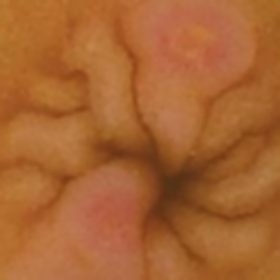} }}%
    \subfloat[\centering 20.7203]{{\includegraphics[width=3.2cm,height=3.2cm]{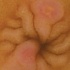} }}%
    \subfloat[\centering 20.7239]{{\includegraphics[width=3.2cm,height=3.2cm]{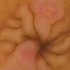} }}\\%
    
    \caption{The experimental justification showing the consistency of the newly proposed EndoQM measure through the degradation pipeline from HR to LR on a sample of WCE Kvasir dataset \citep{kvasir}. Below to each result, we depict value of EndoQM metric. (a) Original image of size $280\times280$, (b) degraded version of (a) with LPF, (c) bicubically downsampled versions of (b) with dimensions $70\times70$ and (d) noisy version of (c) with size $70\times70$.}%
    \label{fig:endo2}%
    
\end{figure*}

\subsection{Usability of Endoscopy Quality Metric (EndoQM)}
To show the usability of domain specific EndoQM measure for endoscopic images, we carried a series of experiments on conventional and capsule images to measure their quantitative performance. The objective of this analysis is to assess whether EndoQM captures distortions in endoscopic images subjected to various degradations unlike the standard no-reference metrics designed for general purpose images and thereby establish its suitability as a domain-specific image quality metric. Fig.~\ref{fig:endo1} and Fig.~\ref{fig:endo2} show the results of these experiments conducted on conventional and capsule images, respectively. In Fig.~\ref{fig:endo1}(a) and Fig.~\ref{fig:endo2}(a), the sample of conventional HR endoscopic and WCE images are displayed along with their EndoQM measure. The original images (i.e., Fig.~\ref{fig:endo1}(a) and Fig.~\ref{fig:endo2}(a)) are passed through a Gaussian Low-Pass Filter (LPF) with kernel size of $5 \times 5$ to attenuate high-frequency components, thereby simulating the degradation of fine texture and edge details commonly observed in low-quality endoscopic images (See Fig.~\ref{fig:endo1}(b) and Fig.~\ref{fig:endo2}(b)). One can see the values of EndoQM measure degrade and thus, ensuring the loss of high-frequency details in terms of quantitative value. Additionally, to show the degradation associated to downsampling operation, the filtered images in Fig.~\ref{fig:endo1}(b) and Fig.~\ref{fig:endo2}(b) are further downsampled by a factor of $ \times 4$, mimicking real-world LR endoscopic observations. The results of these experiments are depicted in Fig.~\ref{fig:endo1}(c) and Fig.~\ref{fig:endo2}(c). By observing the values of EndoQM measure in these cases, it can be deduced that the EndoQM score of these downsampled images are higher, showing downsampling degradation in images. Finally, sensor noise and transmission artifacts are also introduced by adding Gaussian noise with mean $0$ and standard deviation $10$ into Fig.~\ref{fig:endo1}(c) and Fig.~\ref{fig:endo2}(c) which further reduces perceptual quality (see Fig.~\ref{fig:endo1}(d) and Fig.~\ref{fig:endo2}(d)), where values of EndoQM measure are also increasing. Thus, the EndoQM is specifically designed to assess endoscopic image quality, an ideal metric should yield higher degradation scores for images with reduced perceptual fidelity and if EndoQM is a robust evaluator for endoscopic imaging, its score should reflect a progressive decline in image quality across the different degradation levels, i.e., the LPF-processed image, the downsampled image, and the noisy image should exhibit significantly higher EndoQM scores than the original endoscopic images as displayed in Fig.~\ref{fig:endo1}(a) and Fig.~\ref{fig:endo2}(a). 

This experimental validation underscores the effectiveness of EndoQM in capturing clinically relevant quality degradations, distinguishing it from generic reference-less image quality metrics that may not fully align with endoscopic imaging characteristics. By demonstrating its sensitivity to key degradations in endoscopic images, EndoQM establishes itself as a reliable metric for evaluating image quality in real-world medical imaging applications, particularly in WCE and conventional endoscopy-based tasks. Consequently, the EndoQM metric assesses the quality of SR endoscopic images by examining local image statistics and comparing them with the established pristine quality model, thus providing a robust, domain-specific evaluation of image fidelity and perceptual realism. By overcoming the limitations of NIQE in the context of medical imaging, EndoQM delivers a more dependable and precise evaluation framework for endoscopic image quality, ensuring that the improvements brought about by SR techniques meet clinical diagnostic requirements. Therefore, this metric signifies a pivotal element in the assessment of image quality within the realm of medical imaging.

\subsection{Performance Analysis on Newly Derived Kvasir Dataset}
This section presents the qualitative and quantitative analysis of the proposed method on the newly derived Kvasir dataset, comparing its performance against state-of-the-art approaches. The dataset curated from the original Kvasir dataset for SR task consists of $1,000$ LR samples for testing purposes with a resolution of $280 \times 280$ pixels. The SR results obtained using the proposed unsupervised transformer-based SR model are evaluated against the existing unsupervised models, including SRResCGAN \citep{SRResCGAN}, DUSGAN \citep{dusgan}, dSRVAE \citep{dSRVAE}, ZSSR \citep{zssr}, DASR \citep{arr41}, and MDASR \citep{MDASR} for upscaling factor $\times 4$. Notably, all models, except MDASR, were originally trained in an unsupervised manner on natural images, whereas MDASR is specifically designed for training and evaluation on WCE images. Thus, to ensure fair comparison, we re-trained the natural image-trained models (SRResCGAN \citep{SRResCGAN}, DUSGAN \citep{dusgan}, dSRVAE \citep{dSRVAE}, ZSSR \citep{zssr} and DASR \citep{arr41}) on WCE images and conducted a comprehensive qualitative and quantitative evaluation against the proposed method.
\vspace{-0.4cm}
\subsubsection{Qualitative Analysis}
\begin{figure*}[t!]
\centering
    \subfloat[\centering LR \\ (11.5602)]{{\includegraphics[width=3.2cm,height=3.2cm]{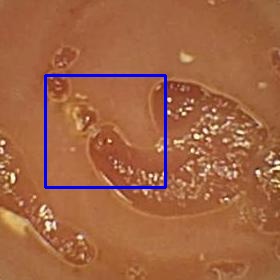} }}%
    \subfloat[\centering SRResCGAN \citep{SRResCGAN} \\ (8.6431)]{{\includegraphics[width=3.2cm,height=3.2cm]{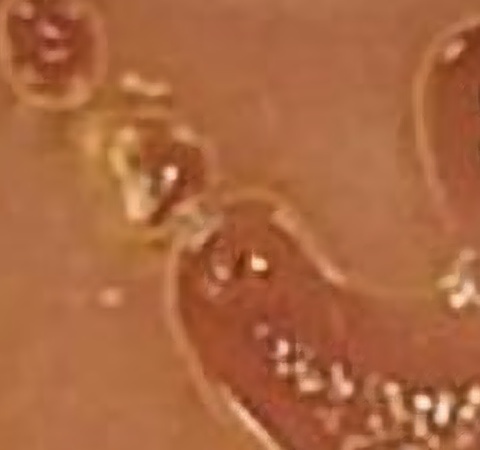} }}%
    \subfloat[\centering DUSGAN \citep{dusgan} \\ (13.6732)]{{\includegraphics[width=3.2cm,height=3.2cm]{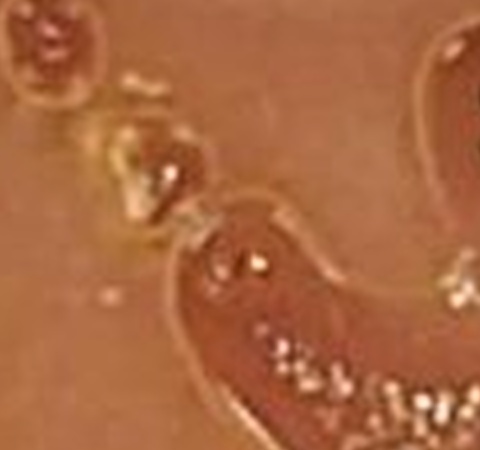} }}%
    \subfloat[\centering dSRVAE \citep{dSRVAE} \\ (9.2218)]{{\includegraphics[width=3.2cm,height=3.2cm]{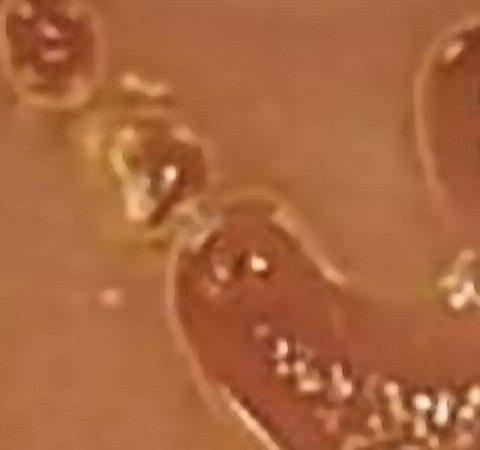} }}\\%
    \vspace{-0.3cm}
    \subfloat[\centering ZSSR \citep{zssr}  \\(12.1514)]{{\includegraphics[width=3.2cm,height=3.2cm]{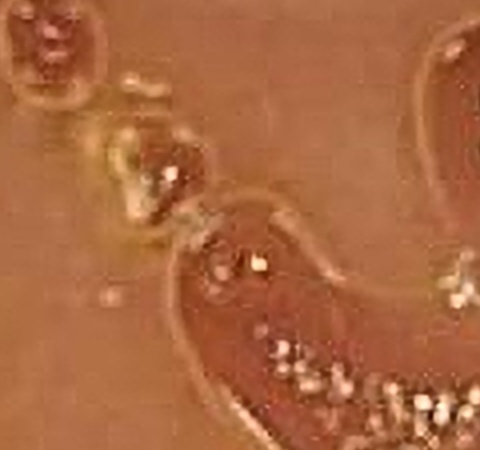} }}%
    \subfloat[\centering DASR \citep{arr41} \\  (14.1511)]{{\includegraphics[width=3.2cm,height=3.2cm]{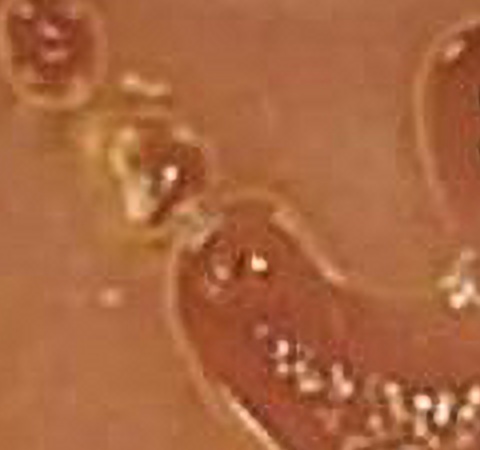} }}%
    \subfloat[\centering MDASR \citep{MDASR} \\  (8.7219)]{{\includegraphics[width=3.2cm,height=3.2cm]{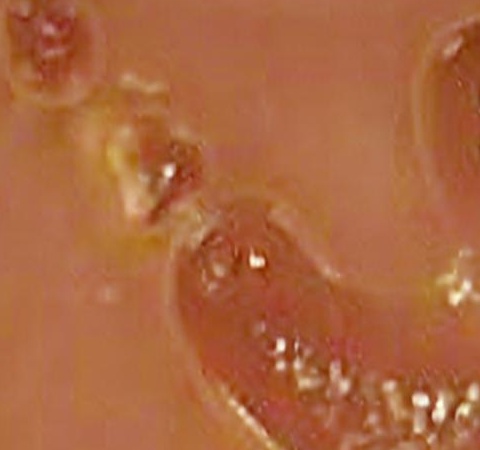} }}%
    \subfloat[\centering Proposed  \\(EndoQM$\downarrow$ = 7.0383)]{{\includegraphics[width=3.2cm,height=3.2cm]{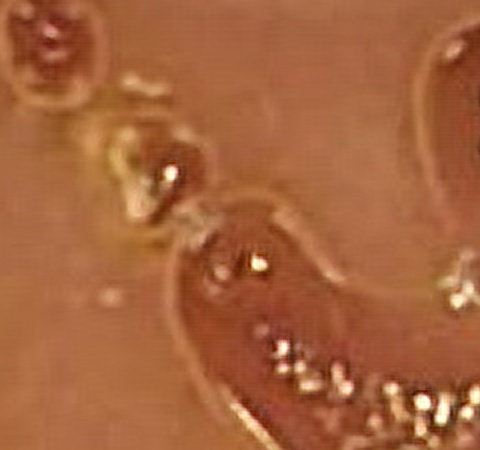} }}%
    \caption{The qualitative comparison of proposed and other existing unsupervised SR methods for upsampling factor $\times4$ on newly edited Kvasir dataset.}%
    \label{fig:f2}%
    \vspace{-0.2cm}
\end{figure*}

The qualitative results demonstrate the ability of the proposed approach to significantly enhance the quality of WCE images, a crucial factor for accurate clinical diagnostics. The Fig.~\ref{fig:f2} showcases a qualitative comparison between the proposed method and six other SR techniques, including SRResCGAN \citep{SRResCGAN}, DUSGAN \citep{dusgan}, dSRVAE \citep{dSRVAE}, ZSSR \citep{zssr}, DASR \citep{arr41}, and MDASR \citep{MDASR}, for the different endoscopic LR images. EndoQM score is noted in Fig.~\ref{fig:f2} to demonstrate the quantitative performance obtained using each SR method\footnote{Due to space constraints, additional figures have been included in the \emph{supplementary material} for further reference.}. One can note that the LR input images (Fig.~\ref{fig:f2}(a)) show severe degradation, including blurring and the loss of structural details, making it challenging to discern the fine features necessary for medical interpretation. This is also supported with the quantitative values of EndoQM measure obtained for LR samples, high values indicate poor preservation of high-frequency details. 

In Fig.~\ref{fig:f2}(b, c), the SR results obtained using SRResCGAN and DUSGAN SR methods are presented where poor reconstruction of details with lack in high-frequency contents can be seen. Similarly, the SR image obtained using dSRVAE SR method in Fig.~\ref{fig:f2}(d) generates smoother outputs, while the SR images obtained using ZSSR and DASR methods (see Fig.~\ref{fig:f2}(e, f)) reconstruct better output compared to the earlier methods, however, they fail to preserve intricate details adequately. The noisy pixels in the reconstruction in the SR images obtained can be seen for ZSSR and DASR methods. The Fig.~\ref{fig:f2}(g) displays the SR image obtained using MDASR SR method which is trained on domain-specific endoscopic images which generates over-smooth image and unrefined high-frequency textures. The SR output obtained using the proposed method (i.e., (\emph{UnCapsTSR}) in Fig.~\ref{fig:f2}(h) demonstrates better clarity and details in reconstruction, preserving fine structural elements and textures making it visually appealing. The high-frequency details associated to the bordering pixels of different parts and regions are preserved well compared to other SR techniques. Further, the value of EndoQM measure for the proposed method also supporting the visual observation while comparing to all other SR methods.
\vspace{-0.4cm}

\subsubsection{Quantitative analysis}

\begin{table*}[!t]
\caption{The quantitative analysis of proposed and other methods. Here, first and second highest values are highlighted with \textcolor{red}{red} and \textcolor{blue}{blue} colors, respectively.}
\resizebox{\textwidth}{!}{%
\Huge
\renewcommand{\arraystretch}{1.25}
\begin{tabular}{lcccc|llll|llll}
\hline \hline
\multicolumn{1}{c}{\multirow{2}{*}{ Method ($\times 4)$}} & \multicolumn{4}{c}{ Newly edited Kvasir Dataset}                    & \multicolumn{4}{c}{ KID dataset } & \multicolumn{4}{c}{ GIANA Dataset } \\ \cline{2-13} 
\multicolumn{1}{c}{}                        &  BRISQUE$\downarrow$ & \multicolumn{1}{l}{ PIQE$\downarrow$} &  NIQE $\downarrow$   &  EndoQM $\downarrow$     &  BRISQUE$\downarrow$   &  PIQE$\downarrow$     &  NIQE$\downarrow$    &  EndoQM $\downarrow$     &  BRISQUE$\downarrow$   &  PIQE$\downarrow$      &  NIQE $\downarrow$    &  EndoQM $\downarrow$     \\ \hline \hline
 SRResCGAN \citep{SRResCGAN}                                  &  66.1341 &  58.1722                  &  \textcolor{red}{4.4847} &  10.3484 &  88.7592   &  75.7123  &  5.1225  &  10.7286  &  86.5817   &  82.1627   &  4.9225   &  9.7258   \\ 
 DUSGAN \citep{dusgan}                                     & 72.1658 & 87.5359                  & 7.3791 & 13.8334 & 59.9603   & 54.9143  & 6.6590  & 17.3653  & 58.0079   & 51.3542   & 5.9432   & 13.4425  \\ 
 dSRVAE \citep{dSRVAE}                                      & 56.4932 & 64.9396                  & 4.8916 & 10.8352 & \textcolor{blue}{56.6771}   & 47.8164  & 5.9949  & 13.5611  & \textcolor{blue}{46.6371}   & \textcolor{blue}{43.2417}   & 5.6909   & 10.0163  \\ 
 DASR\citep{arr41}                                       & 63.7744 & 91.8829                  & 6.0622 & 14.2492 & 71.3107   & \textcolor{blue}{42.6297}  & 5.4863  & 13.5945  & 63.1576   & 53.8924   & 4.2092   & 10.0339  \\ 
 ZSSR \citep{zssr}                                       & 64.7164 & 85.2007                  & 6.0950 & 13.1060 & 83.7318   & 81.1143  & 4.9983  & 11.3095  & 86.9541   & 81.2241   & 5.0458   & 8.4063   \\ 
 MDASR \citep{MDASR}                                       & \textcolor{blue}{54.9760} & \textcolor{blue}{55.9510}                  & 4.9506 & \textcolor{blue}{8.9283}  & 76.7706   & 47.4550  & \textcolor{blue}{4.8297}  & \textcolor{blue}{7.6567}   & 71.0476   & 68.4324   & \textcolor{blue}{4.8809}   & \textcolor{blue}{7.7718}   \\ 
 Proposed                                    & \textcolor{red}{42.0358} & \textcolor{red}{41.8822}                  & \textcolor{blue}{4.5346} & \textcolor{red}{7.0914}  & \textcolor{red}{44.5833}   & \textcolor{red}{29.3739}  & \textcolor{red}{4.0198}  & \textcolor{red}{6.7696}   & \textcolor{red}{53.2019}   & \textcolor{red}{29.7511}   & \textcolor{red}{3.7547}   & \textcolor{red}{6.9723}   \\ \hline
\hline
\end{tabular}%
}
\label{table1}
\vspace{-0.5cm}
\end{table*}

The SR results obtained using the proposed and other SR methods are quantitatively evaluated in terms of no-reference image quality metrics such as BRISQUE, NIQE, PIQE, and the newly introduced EndoQM in Table~\ref{table1}. These metrics provide insights into the perceptual quality of the super-resolved WCE images without relying on ground-truth HR images. The lower values of BRISQUE, PIQE, NIQE and EndoQM scores indicate better quality in the SR observations. In Table~\ref{table1}, the top two values are highlighted with red and blue colored texts, respectively, for the convenience of the reader. It can be deduced that the proposed model achieves the lowest BRISQUE score among all methods, indicating the highest quality and minimal artifacts. The PIQE evaluates image quality based on local distortion detection and the proposed model attains lowest scores among all, indicating minimal perceptual distortions and superior visual quality compared to the other SR methods. The SR methods such as ZSSR and DASR perform reasonably well in qualitative results despite introducing minor distortions in a few regions as captured by their higher PIQE scores. 
Further, the quantitative score in terms of NIQE measure is also best using the proposed model than the other competing SR methods, demonstrating its ability to restore perceptually natural textures in WCE images. However, the performance gap highlights the limitations of NIQE when applied to medical images. Finally, the EndoQM metric, specifically tailored for endoscopic images, provide the most domain-relevant evaluation. The proposed model consistently outperforms all other methods, achieving the lowest EndoQM scores highlighting superior ability of the proposed model to reconstruct high-frequency details and preserving structural and textural integrity, crucial for endoscopic diagnostics. Overall, the proposed model achieves the better quantitative performance across all metrics, along with newly proposed EndoQM results validating its suitability for clinical applications by aligning closely with the domain-specific requirements of endoscopic imaging for newly curated Kvasir dataset.

\vspace{-0.3cm}
\subsection{Performance analysis on KID dataset}
Further, to evaluate the generalization capability of the proposed model, additional experiments are conducted on the KID dataset \citep{KID}, an external dataset containing endoscopic images with distinct characteristics (Details in Table~\ref{dataset}). It is important to note here that the images of this dataset are not presented in the training of any algorithm.

\subsubsection{Qualitative Analysis}

\begin{figure*}[t!]
    \centering
    \subfloat[\centering LR \\ (12.9284)]{{\includegraphics[width=3.2cm,height=3.2cm]{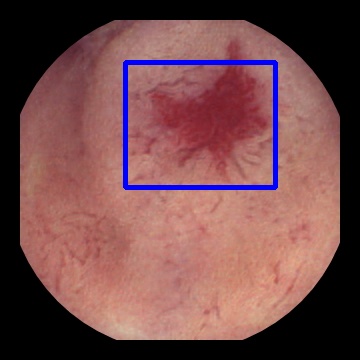} }}%
    \subfloat[\centering SRResCGAN \citep{SRResCGAN} \\ (10.0397)]{{\includegraphics[width=3.2cm,height=3.2cm]{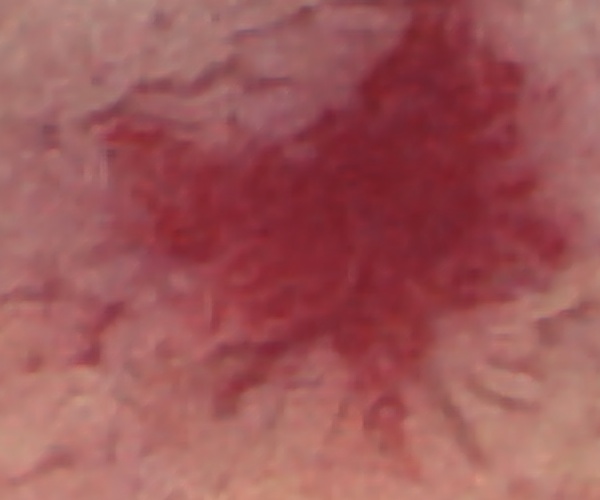} }}%
    \subfloat[\centering DUSGAN \citep{dusgan} \\ (20.0316)]{{\includegraphics[width=3.2cm,height=3.2cm]{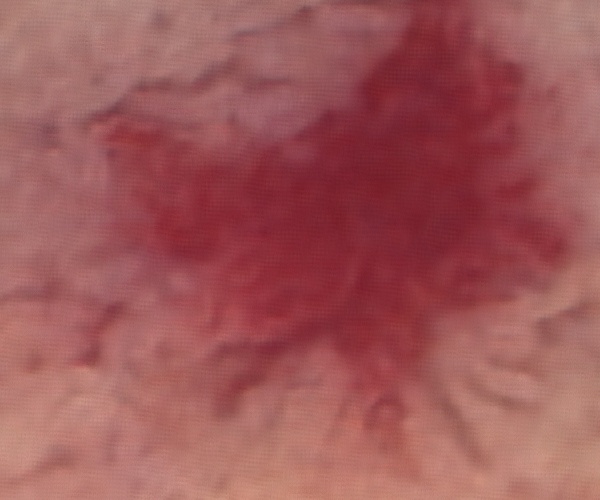} }}%
    \subfloat[\centering dSRVAE \citep{dSRVAE} \\ (16.1445)]{{\includegraphics[width=3.2cm,height=3.2cm]{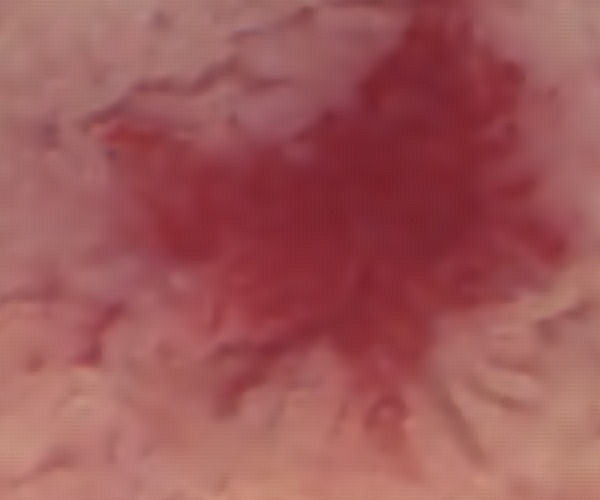} }}\\
    \vspace{-0.3cm}
    \subfloat[\centering ZSSR \citep{zssr} \\ (11.8243)]{{\includegraphics[width=3.2cm,height=3.2cm]{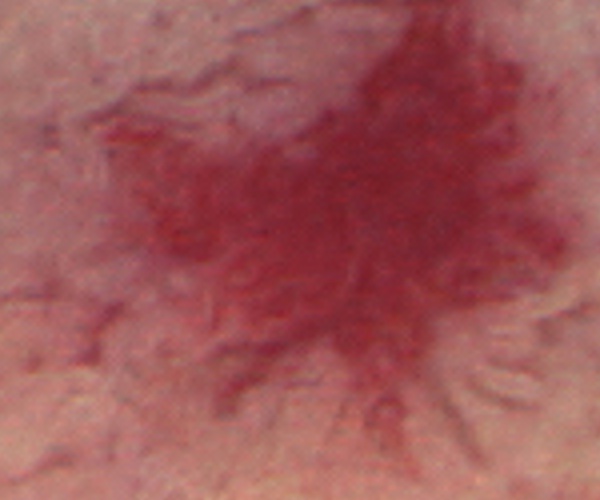} }}%
    \subfloat[\centering DASR \citep{arr41}\\  (13.6794)]{{\includegraphics[width=3.2cm,height=3.2cm]{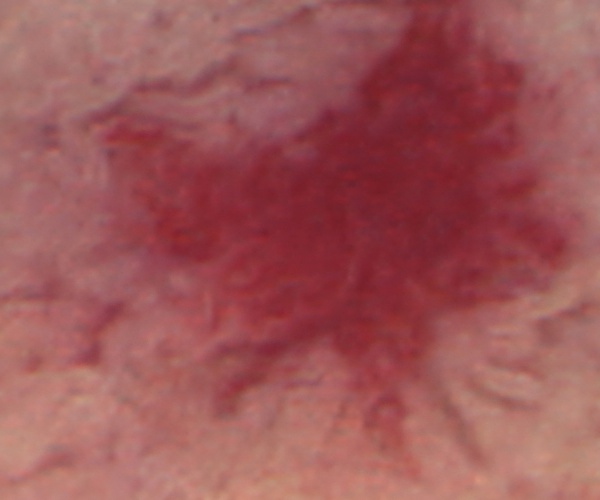} }}%
    \subfloat[\centering MDASR \citep{MDASR} \\(7.6872)]{{\includegraphics[width=3.2cm,height=3.2cm]{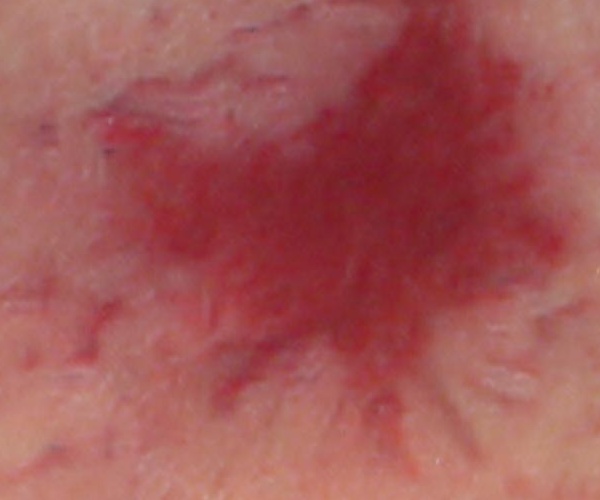} }}%
    \subfloat[\centering Proposed \\ (EndoQM $\downarrow$ = 6.7876)]{{\includegraphics[width=3.2cm,height=3.2cm]{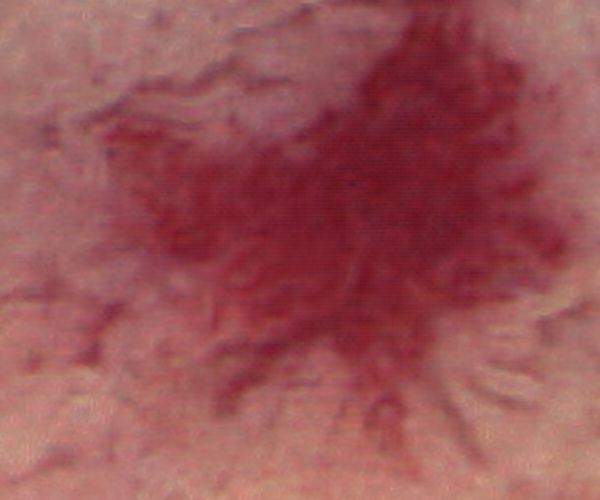} }}%
    \caption{The qualitative comparison of proposed and other existing unsupervised SR methods for upsampling factor $\times4$  on KID Dataset.}%
    \label{kid1}%
   
\end{figure*}

The qualitative comparison of the proposed model with state-of-the-art methods on the KID dataset for an upscaling factor of $\times 4$ is shown in Fig.~\ref{kid1}\footnote{Due to space constraints, additional figures on KID have been included in the \emph{supplementary material}.}. The LR images in Fig.~\ref{kid1}(a) exhibit significant degradation with blurred lesion boundaries and loss of high-frequency details. 
Existing SR methods like SRResCGAN and DUSGAN (see Fig.~\ref{kid1}(b, c)) enhance lesion shape and texture clarity to some extent but often introduce artifacts or over-smoothing, obscuring vascular details. Similarly, the dSRVAE method (see Fig.~\ref{kid1}(d)) produces a smoother SR image but blurs the red region. The DASR and ZSSR methods (see Fig.~\ref{kid1}(e, f)) improve visibility in broader lesion areas yet struggle to reconstruct intricate features like capillary networks. While MDASR (see Fig.~\ref{kid1}(g)) generates a clean and smooth output, it over-smooths critical regions, leading to a loss of capillary networks and textural details needed for early-stage lesion identification and complex vascular pattern analysis. In contrast, the proposed model (see Fig.~\ref{kid1}(h)) excels in preserving vascular details and reconstructing lesion textures, minimizing artifacts and enabling detailed clinical diagnostics.

\vspace{-0.4cm}
\subsubsection{Quantitative Analysis}

The quantitative analysis to evaluate the generalization performance of the proposed model on the KID dataset is displayed in Table~\ref{table1}. Here, the proposed model demonstrates significant improvements across all metrics compared to other state-of-the-art methods. The SRResCGAN and ZSSR SR methods obtain high PIQE and BRISQUE scores, reflecting their inability to recover fine details as they tend to generate noisy artifacts in SR output. The MDASR shows moderate improvements leveraging its endoscopy-specific training, yet it fails to fully restore high-frequency details. In contrast, the proposed model attains the lowest (better) PIQE and BRISQUE scores, indicating minimal perceptual distortions and superior visual quality with restored structural coherence. Similarly, the NIQE scores further validated the perceptual quality of the proposed model, with significantly better results. However, the most notable performance is observed in the EndoQM metric, specifically tailored for endoscopic images. The proposed model obtains lowest EndoQM scores, reflecting its ability to effectively restore subtle textures, intricate vascular patterns, and diagnostically relevant details crucial for clinical applications. This superior performance of the proposed method can be attributed to the model's transformer-based architecture and its domain adaptation strategies, which enable it to capture long-range dependencies and align with the unique statistical properties of endoscopic images.

\subsection{Performance analysis on GIANA dataset}
Finally, the performance of the proposed model is also evaluated on the GIANA \citep{GIANA} dataset (Refer Table~\ref{dataset}) to assess its generalization capability beyond the original training set.

\begin{figure*}[t!]
    \centering
    \subfloat[\centering LR \\ (11.8082)]{{\includegraphics[width=3.2cm,height=3.2cm]{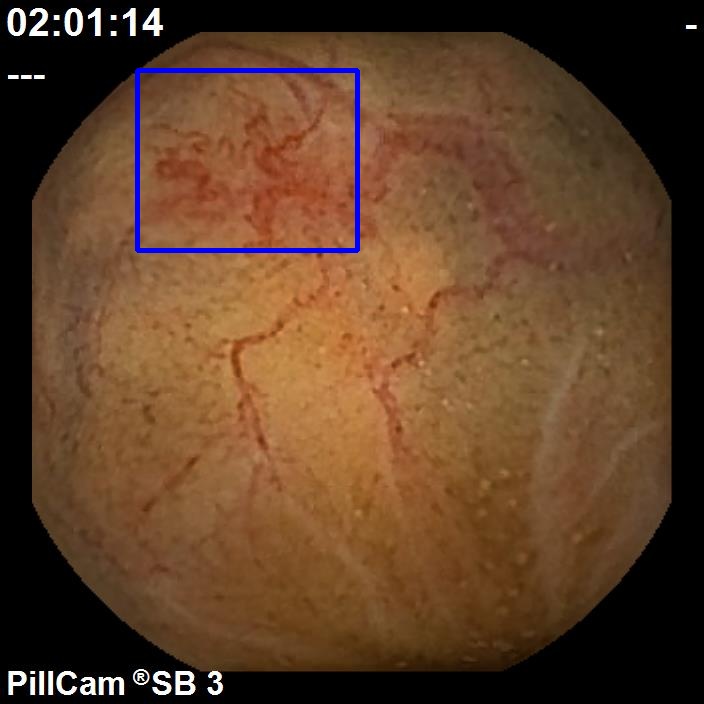} }}%
    \subfloat[\centering SRResCGAN \citep{SRResCGAN} \\ (10.4429)]{{\includegraphics[width=3.2cm,height=3.2cm]{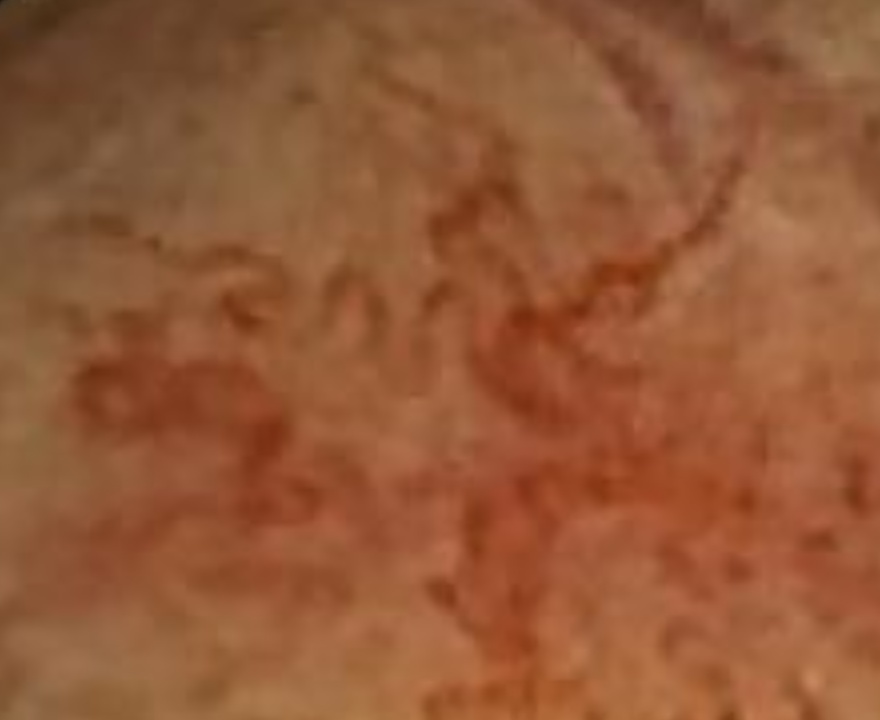} }}%
    \subfloat[\centering DUSGAN \citep{dusgan} \\ (13.4488)]{{\includegraphics[width=3.2cm,height=3.2cm]{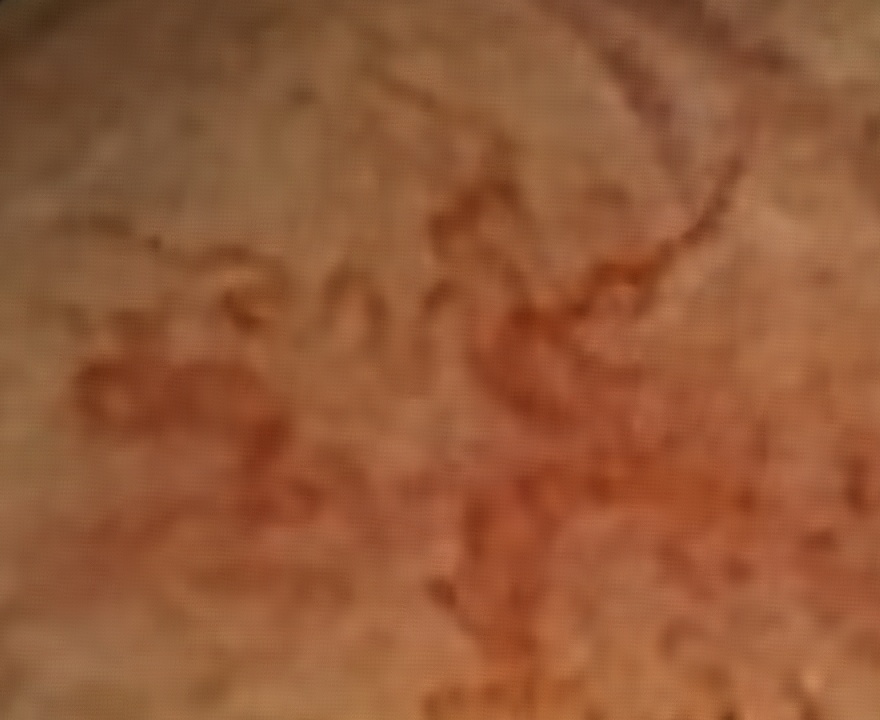} }}%
    \subfloat[\centering dSRVAE \citep{dSRVAE} \\ (8.5677)]{{\includegraphics[width=3.2cm,height=3.2cm]{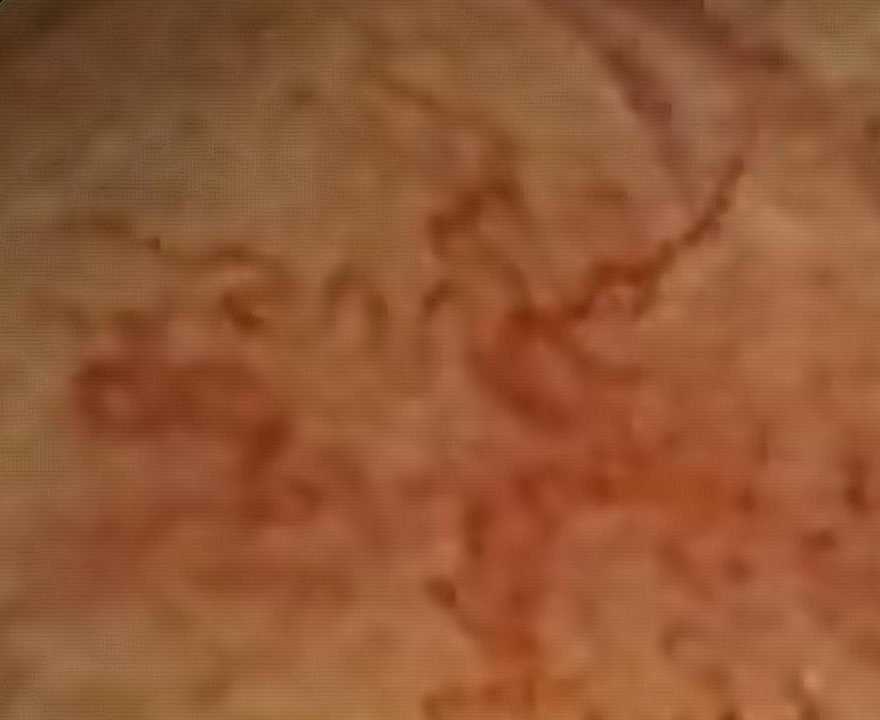} }}\\
    \vspace{-0.3cm}
    \subfloat[\centering ZSSR \citep{zssr} \\  (9.6327)]{{\includegraphics[width=3.2cm,height=3.2cm]{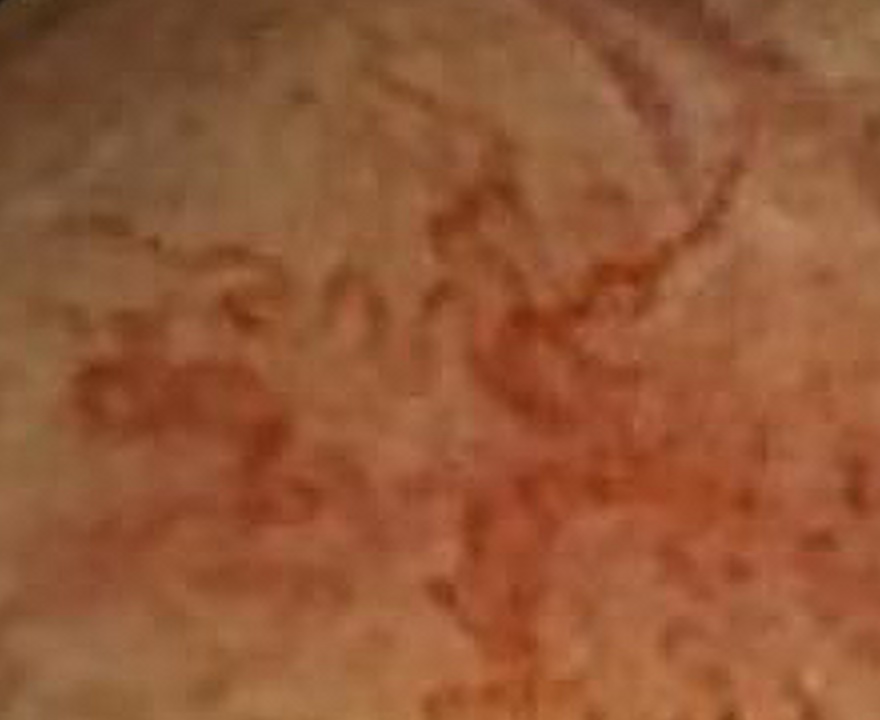} }}%
    \subfloat[\centering DASR \citep{arr41} \\  (8.6178)]{{\includegraphics[width=3.2cm,height=3.2cm]{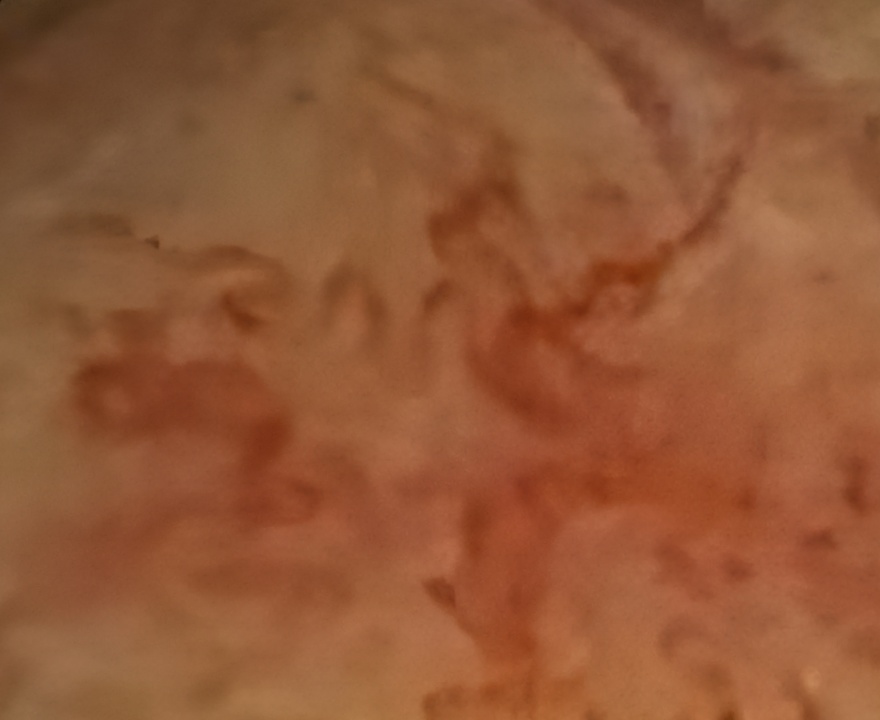} }}%
    \subfloat[\centering MDASR \citep{MDASR} \\ (7.1949)]{{\includegraphics[width=3.2cm,height=3.2cm]{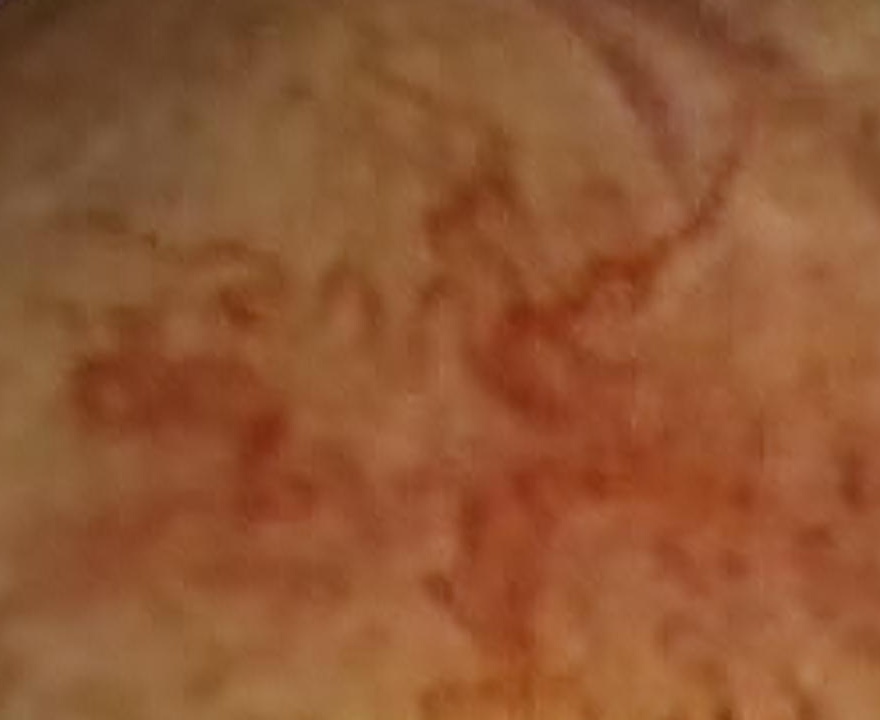} }}%
    \subfloat[\centering Proposed  \\(EndoQM$\downarrow$ = 6.5410)]{{\includegraphics[width=3.2cm,height=3.2cm]{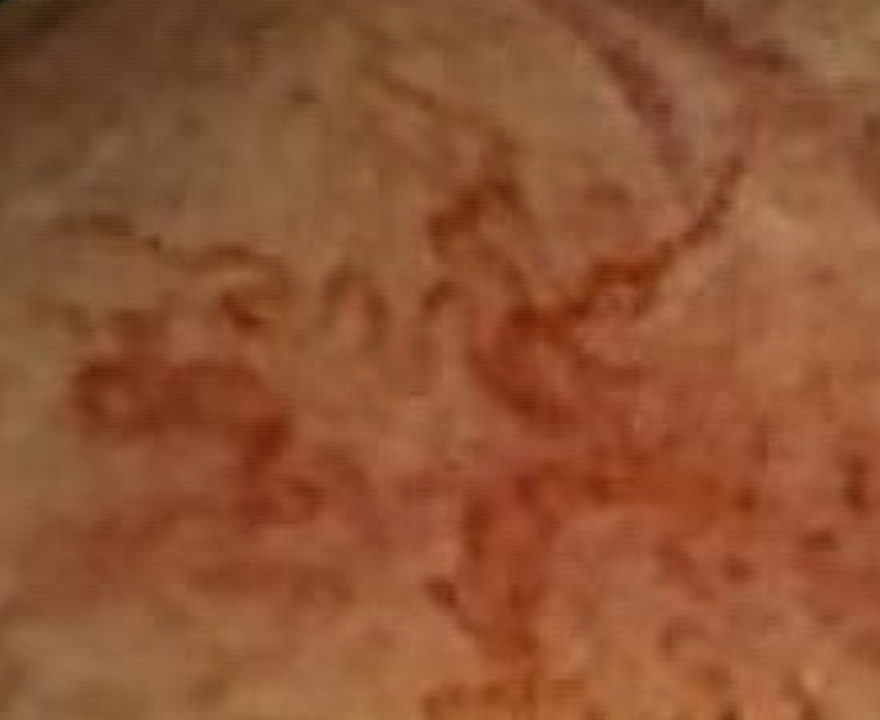} }}%
    \caption{The qualitative comparison of proposed and other existing unsupervised SR methods for upsampling factor $\times4$ on GIANA Dataset.}%
    \label{giana1}%
    \vspace{-0.4cm}
\end{figure*}

\vspace{-0.3cm}
\subsubsection{Qualitative Analysis}
The visual SR results of the proposed model are compared with other state-of-the-art models in Fig.~\ref{giana1}\footnote{Due to space constraints, additional figures on GIANA have been included in the \emph{supplementary material}.}. Here, the LR input image exhibits significant blurring and loss of fine structures. The SR images generated by SRResCGAN and DUSGAN (see Fig.~\ref{giana1}(b, c)) enhance the lesion area but fail to preserve vein information crucial for accurate diagnosis. Similarly, the ZSSR method (see Fig.~\ref{giana1}(f)) provides moderate lesion sharpness but lacks sufficient texture detail for clinical reliability. The dSRVAE and MDASR methods (see Fig.~\ref{giana1}(d, g)) offer improved lesion boundary sharpness but suffer from over-smoothing, resulting in texture loss. In contrast, the proposed model (see Fig.~\ref{giana1}(h)) significantly outperforms other methods by reconstructing sharper lesion boundaries while preserving natural texture patterns, delivering enhanced clarity and diagnostic reliability suitable for clinical applications.

\subsubsection{Quantitative Analysis}
Table~\ref{table1} depicts the quantitative analysis on GIANA dataset with the other state-of-the-art models. By visualizing this table, one can note that the results of the proposed model are significant than the other existing methods across all evaluated metrics. For instance, the SR models such as SRResCGAN \citep{SRResCGAN} and ZSSR \citep{zssr} exhibit high PIQE and BRISQUE scores, highlighting their limitations in recovering fine details and their propensity to generate outputs with noticeable noisy artifacts. While MDASR demonstrates moderate improvements by leveraging endoscopy-specific training, it fails to fully reconstruct high-frequency details and thus,  introduces minor distortions in regions with complex textures. In contrast, the proposed model achieves the lowest (i.e., optimal) PIQE and BRISQUE scores, signifying minimal perceptual distortions and superior structural fidelity in its reconstructions. Additionally, the NIQE scores further corroborated the model's enhanced perceptual quality, showcasing a marked improvement over the competing SR methods. Notably, the proposed model excels in the EndoQM metric. Quantitative metric reflects an exceptional ability to recover subtle textures, intricate patterns, and diagnostically critical details.

\subsection{Statistical Analysis}

To assess the reliability of the quantitative results obtained using the proposed (i.e. \emph{UnCapsTSR}) and state-of-the-art SR methods, we conducted a statistical evaluation employing the Analysis of Variance (ANOVA) test. In this test, the quantitative results for BRISQUE, PIQE, NIQE, and EndoQM metrics computed at a $95\%$ Confidence Interval (CI) are presented in Table II of \emph{supplementary material}, while the corresponding box-plots illustrating the distribution of these quality metrics across different SR methods on the newly curated Kvasir Capsule test dataset are depicted in Fig. 9 of \emph{supplementary material}. To further strengthen the reliability and robustness of the proposed \emph{UnCapsTSR} framework, we performed a statistical validation analysis using a two-tailed Z-test \citep{z1,z4}. The purpose of this test is to determine whether the performance improvements observed with \emph{UnCapsTSR} over baseline models are statistically significant or could be attributed to random variation. The distribution plots for the Z-tests across all baselines and metrics is shown in Fig. 10 in \emph{supplementary material}. To further establish the robustness of the proposed \emph{UnCapsTSR} framework, we have conducted a Kolmogorov–Smirnov (K–S) statistical analysis \citep{z2,z3} on the no-reference quality assessment scores of WCE images. The K–S test is a non-parametric method that measures the maximum vertical distance (D) between two Cumulative Distribution Functions (CDFs), thereby quantifying whether two samples are drawn from the same distribution. In this study, we applied two-sample K–S tests between the \emph{UnCapsTSR} model and each baseline method across non reference metrics: BRISQUE, PIQE, NIQE, and EndoQM. The Empirical Cumulative Distribution Functions (ECDFs) of the scores for all models is shown in Fig. 11 in \emph{supplementary material}.
\vspace{-0.4cm}

\subsection{Ablation Study}
To comprehensively analyze the effectiveness of the proposed model (i.e., \emph{UnCapsTSR}), an ablation study is carried out. This study evaluates the impact of different loss functions and network configurations on the performance across multiple datasets, including the newly edited Kvasir, KID, and GIANA datasets. The evaluation is performed using different reference-less quality metrics: BRISQUE, PIQE, NIQE, and EndoQM which are summarized in Table III in the \emph{supplementary material}. The SR results obtained using these experiments are also depicted in Fig. 12 and Fig. 13 in the \emph{supplementary material} to see their performance in visual aspects.  
\vspace{-0.4cm}

\section{Conclusion}
\label{sec:conwce}
The WCE is a preferred over traditional endoscope in the medical field to inspect patient's GI track. Despite many benefits of this technology, its hardware limitations lead to sense data with coarser resolution. This work presented an approach named as \emph{UnCapsTSR} to enhance the spatial resolution of the LR WCE data using unsupervised training eliminating the need for  genuine LR-HR paired datasets for supervised training. The approach makes use of transformer in GAN framework with direct domain transfer from LR to SR without the explicit use of degradation learning. Along with perceptual GAN loss, the \emph{UnCapsTSR} also employs the novel Bilateral Total Variation (BTV) loss to improve the spatial continuity in the final SR image. To check the generalizability of the proposed method, it has experimented on the number of other datasets whose samples are not part of the training and their SR results are checked with many unsupervised SR approaches. Through this study, we also introduced domain-specific quantitative reference-less index designed to gauge to quality of WCE SR models. Additionally, we also introduced the curated version of the Kvasir dataset to train different SR algorithms. The visual and quantitative analysis of the proposed method demonstrates its effectively over the other state-of-the-art SR methods alongside the statistical evaluation in terms of ANOVA test, Z-test and Kolmogorov–Smirnov (K–S) test. Future works can also extend the proposed approach for super resolution by taking into account pathology specific details to have better quality images while improving the diagnostic quality.
\vspace{-0.5cm}
\section*{Acknowledgements}
The authors are thankful to the CapsNetwork – International Network for Capsule Imaging in Endoscopy (project no. 322600) funded by the Research Council of Norway.

\vspace{-0.5cm}
\bibliographystyle{elsarticle-harv} 
\bibliography{Reference}






\end{document}